\documentclass[10pt]{article}

\usepackage[final]{neurips_2022}

\usepackage{enumitem}

\usepackage[utf8]{inputenc} 
\usepackage[T1]{fontenc}    
\usepackage[colorlinks,citecolor=blue]{hyperref}
\usepackage{url}            
\usepackage{booktabs}       
\usepackage{amsfonts}       
\usepackage{nicefrac}       
\usepackage{microtype}      
\usepackage{xcolor}         

\usepackage{wrapfig}
\usepackage{amsthm,verbatim,amssymb,amsfonts,amscd,graphicx}
\theoremstyle{plain}
\newtheorem{theorem}{Theorem}

\newtheorem{lemma}{Lemma}
\newtheorem*{remark}{Remark}
\newtheorem{proposition}{Proposition}

\newtheorem{definition}{Definition}

\usepackage[bottom]{footmisc}
\usepackage{caption}
\usepackage{subcaption}
\usepackage{helvet}  
\usepackage{courier}  
\usepackage{url}  
\usepackage{graphicx}  
\usepackage{multirow}
\usepackage{amsthm}
\usepackage{color}
\usepackage{MnSymbol}
\usepackage{makecell}
\usepackage{arydshln}
\usepackage{amsmath}
\usepackage{caption} 
\usepackage{natbib}
\usepackage{bbm}
\usepackage{textcomp}
\usepackage{wrapfig}
\usepackage{algorithm}
\usepackage{algorithmic}

\usepackage{csquotes}

\newcommand{\E}{\mathbb{E}}
\newcommand{\V}{{\rm Var}}
\newcommand{\diag}{{\rm diag}}
\newcommand{\sgn}{{\rm sgn}}

\newcommand{\revise}[1]{{\color{black}{#1}}}

\begin{document}

\title{Exact Solutions of a Deep Linear Network}
\author{Liu Ziyin$^1$, Botao Li$^2$, Xiangming Meng$^3$\\
$^1$Department of Physics, The University of Tokyo\\
$^2$Laboratoire de Physique de l’Ecole normale sup\'erieure, ENS,
Universit\'e PSL,\\CNRS, Sorbonne Universit\'e, Universit\'e de Paris Cit\'e, Paris, France\\
$^3$Institute for Physics of Intelligence, Graduate School of Science, The University of Tokyo
}
\maketitle

\begin{abstract}%
  This work finds the analytical expression of the global minima of a deep linear network with weight decay and stochastic neurons, a fundamental model for understanding the landscape of neural networks. Our result implies that the origin is a special point in deep neural network loss landscape where highly nonlinear phenomenon emerges. We show that weight decay strongly interacts with the model architecture and can create bad minima at zero in a network with more than $1$ hidden layer, qualitatively different from a network with only $1$ hidden layer. Practically, our result implies that common deep learning initialization methods are insufficient to ease the optimization of neural networks in general. 
\end{abstract}


\section{Introduction}
Applications of neural networks have achieved great success in various fields. One central open question is why neural networks, being nonlinear and containing many saddle points and local minima, can sometimes be optimized easily \citep{choromanska2015loss} while becoming difficult and requiring many tricks to train in some other scenarios \citep{glorot2010understanding, gotmare2018closer}. One established approach is to study the landscape of deep linear nets \citep{choromanska2015open}, which are believed to approximate the landscape of a nonlinear net well. A series of works proved the famous results that for a deep linear net, all local minima are global \citep{kawaguchi2016deep, lu2017depth, laurent2018deep}, which is regarded to have successfully explained why deep neural networks are so easy to train because it implies that initialization in any attractive basin can reach the global minimum without much effort \citep{kawaguchi2016deep}. However, the theoretical problem of when and why neural networks can be hard to train is understudied.

In this work, we theoretically study a deep linear net with weight decay and stochastic neurons, whose loss function takes the following form in general:
\begin{equation}\label{eq: intro loss}
    \underbrace{\E_x \E_{\epsilon^{(1)}, \epsilon^{(2)},...,\epsilon^{(D)}} \left(\sum_{i,i_1,i_2,...,i_D}^{d,d_1,d_2,...d_D} U_{i_D} \epsilon^{(D)}_{i_{D}}...\epsilon_{i_2}^{(2)} W_{i_2i_1}^{(2)}\epsilon_{i_1}^{(1)} W_{i_1i}^{(1)} x_i  - y\right)^2}_{L_0} + \underbrace{\gamma_u||U||_2^2 + \sum_{i=1}^D\gamma_i ||W^{(i)}||_F^2}_{ L_2\ reg.},
\end{equation}
where $\E_x$ denotes the expectation over the training set, $U$ and $W^{(i)}$ are the model parameters, $D$ is the depth of the network,\footnote{In this work, we use ``depth" to refer to the number of hidden layers. For example, a linear regressor has depth $0$.} $\epsilon$ is the noise in the hidden layer (e.g., due to dropout), $d_i$ is the width of the $i$-th layer, and $\gamma$ is the strength of the weight decay. Previous works have studied special cases of this loss function. For example, \cite{kawaguchi2016deep} and \cite{lu2017depth} study the landscape of $L_0$ when $\epsilon$ is a constant (namely, when there is no noise). \cite{mehta2021loss} studies $L_0$ with (a more complicated type of) weight decay but without stochasticity and proved that all the stationary points are isolated. Another line of works studies $L_0$ when the noise is caused by dropout \citep{mianjy2019dropout, cavazza2018dropout}. Our setting is more general than the previous works in two respects. First, apart from the mean square error (MSE) loss $L_0$, an $L_2$ regularization term (weight decay) with arbitrary strength is included; second, the noise $\epsilon$ is arbitrary. Thus, our setting is arguably closer to the actual deep learning practice, where the injection of noises to latent layers is common, and the use of weight decay is virtually ubiquitous \citep{krogh1992simple, loshchilov2017decoupled}. One major limitation of our work is that we assume the label $y$ to be $1$-dimensional, and it can be an important future problem to prove whether an exact solution exists or not when $y$ is high-dimensional.

\begin{figure}[t!]
    \centering
    \vspace{-1em}
    \includegraphics[width=0.45\linewidth]{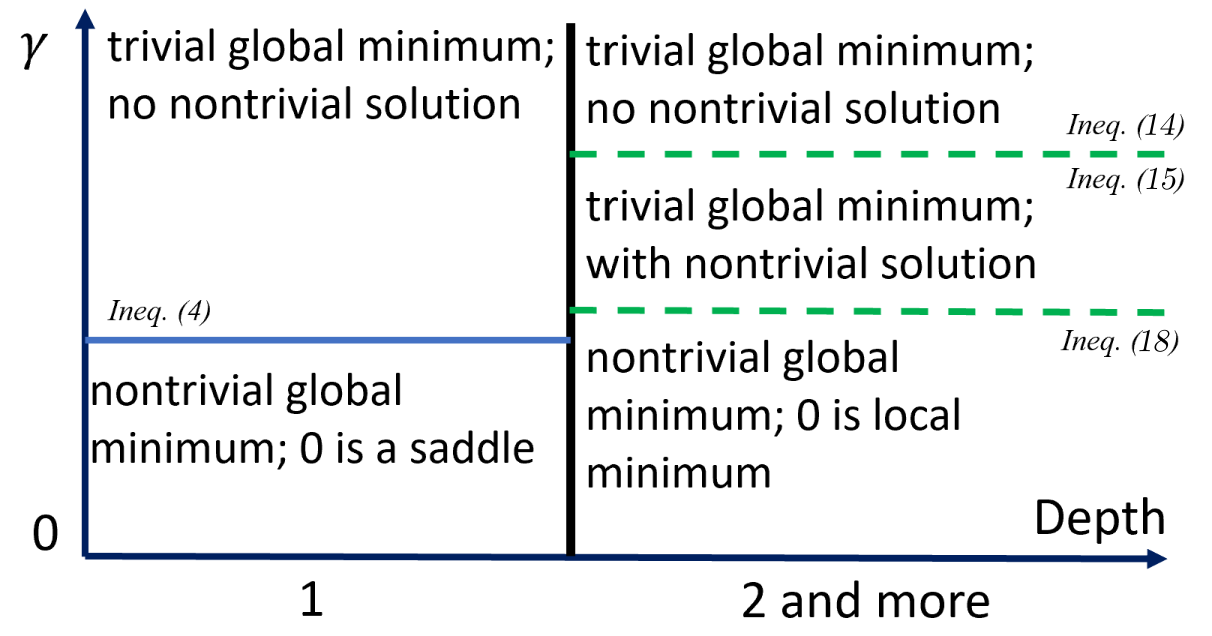}
    \includegraphics[width=0.3\linewidth]{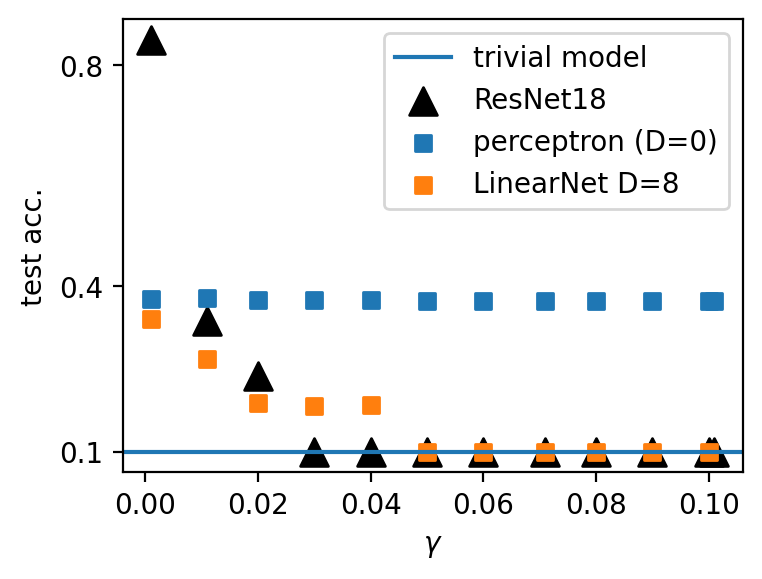}

    \vspace{-0.5em}
    \caption{\small \textbf{Left}: A summary of the network landscape that is implied by the main results of this work when one increases the weight decay strength $\gamma$ while fixing other terms. We show that the landscape of a depth-$1$ net can be precisely divided into two regimes, while, for $D\geq 2$, there exists at least three regimes. The solid blue line indicates that the division of the regimes is precisely understood. The dashed lines indicate that the conditions we found are not tight and may be improved in the future. \textbf{Right}: \color{black} ResNet18 on CIFAR10. The performance of a linear regressor never drops to that of a trivial model, whereas the performance of ResNet18 drops to the level of a trivial model, like a deep linear net with similar depth.}
    \label{fig:summary}
    \vspace{-1em}
\end{figure}

Our foremost contribution is to prove that all the global minimum of an arbitrarily deep and wide linear net takes a simple analytical form. In other words, we identify in closed form the global minima of Eq.~\eqref{eq: intro loss} up to a single scalar, whose analytical expression does not exist in general. We then show that it has nontrivial properties that can explain many phenomena in deep learning. In particular, the implications of our result include (but are not limited to):
\begin{enumerate}
    \item Weight decay makes the landscape of neural nets more complicated;
    \begin{itemize}
            \item we show that bad minima\footnote{Unless otherwise specified, we use the word ``bad minimum" to mean a local minimum that is not a global minimum.} emerge as weight decay is applied, whereas there is no bad minimum when there is no weight decay. This highlights the need to escape bad local minima in deep learning with weight decay.
    \end{itemize}
    \item Deeper nets are harder to optimize than shallower ones;
        \begin{itemize}
            \item we show that a $D\geq 2$ linear net contains a bad minimum at zero, whereas a $D=1$ net does not. This partially explains why deep networks are much harder to optimize than shallower ones in deep learning practice.
        \end{itemize}
    \item Depending on the task, the common initialization methods (such as the Kaiming init.) can initialize a deep model in the basin of attraction of the bad minimum at zero;
    \begin{itemize}
        \item common initialization methods initialize the models at a radius of roughly $1/\sqrt{width}$ around the origin; however, we show that the width of the bad minimum is task-dependent and can be larger than the initialization radius for tasks with a small margin ($||\E[xy]||$);
    \end{itemize}
    \item Thus, the use of (effective) weight decay is a major cause of various types of collapses in deep learning (for example, see Figure~\ref{fig:summary}).
\end{enumerate}

\textbf{Organization}: In the next section, we discuss the related works. In Section~\ref{sec: two layer net}, we derive the exact solution for a two-layer net. Section~\ref{sec: arbitrary depth net} extends the result to an arbitrary depth. In Section~\ref{sec: implications}, we study and discuss the relevance of our results to many commonly encountered problems in deep learning. The last section concludes the work and discusses unresolved open problems. All proofs are delayed to Section~\ref{app sec: proof}. Moreover, additional theoretical results on the effect of including a bias term is considered in Section~\ref{app sec:bias}.  

\textbf{Notation}. For a matrix $W$, we use $W_{i:}$ to denote the $i$-th row vector of $W$. $||Z||$ denotes the $L_2$ norm if $Z$ is a vector and the Frobenius norm if $Z$ is a matrix. The notation $*$ signals an optimized quantity. Additionally, we use the superscript $^*$ and subscript $_*$ interchangeably, whichever leads to a simpler expression. For example, $b_*^2$ and $(b^*)^2$ denote the same quantity, while the former is ``simpler."

\vspace{-2mm}
\section{Related Works}
\vspace{-1mm}


In many ways, linear networks have been used to help understand nonlinear networks. For example, even at depth $0$, where the linear net is just a linear regressor, linear nets are shown to be relevant for understanding the generalization behavior of modern overparametrized networks \citep{hastie2019surprises}. \cite{saxe2013exact} studies the training dynamics of a depth-$1$ network and uses it to understand the dynamics of learning of nonlinear networks. These networks are the same as a linear regression model in terms of expressivity. However, the loss landscape is highly complicated due to the existence of more than one layer, and linear nets are widely believed to approximate the loss landscape of a nonlinear net \citep{kawaguchi2016deep,hardt2016identity,laurent2018deep}. In particular, the landscape of linear nets has been studied as early as 1989 in \cite{baldi1989neural}, which proposed the well-known conjecture that all local minima of a deep linear net are global. This conjecture is first proved in \cite{kawaguchi2016deep}, and extended to other loss functions and deeper depths in \cite{lu2017depth} and \cite{laurent2018deep}. Many relevant contemporary deep learning problems can be understood with deep linear models. For example, two-layer linear VAE models are used to understand the cause of the posterior collapse problem \citep{lucas2019dont, wang2022posterior}. Deep linear nets are also used to understand the neural collapse problem in contrastive learning \citep{tian2022deep}. We also provide more empirical evidence in Section~\ref{sec: implications}.

\vspace{-2mm}
\section{Two-layer Linear Net}\label{sec: two layer net}
\vspace{-1mm}

This section finds the global minima of a two-layer linear net. The data point is a $d$-dimensional vector $x\in \mathbb{R}^{ d}$ drawn from an arbitrary distribution, and the labels are generated through an arbitrary function $y=y(x)\in \mathbb{R}$. For generality, we let different layers have different strengths of weight decay even though they often take the same value in practice. We want to minimize the following objective:
\begin{equation}\label{eq: main loss two layer}
    L_{d, d_1}(U,W) =  \E_x \E_{\epsilon} \left(\sum_j^{d_1} U_j \epsilon_j \sum_i^d W_{ji} x_i  - y\right)^2 + \gamma_w ||W||^2 + \gamma_u ||U||^2,
\end{equation}
where $d_1$ is the width of the hidden layer and $\epsilon_i$ are independent random variables. $\gamma_w > 0$ and $\gamma_u>0$ are the weight decay parameters. Here, we consider a general type of noise with $\E[\epsilon_i] = 1$ and $\E[\epsilon_i\epsilon_j] = \delta_{ij}\sigma^2 + 1$ where $\delta_{ij}$ is the Kronecker's delta, and $\sigma^2 > 0$.\footnote{While we formally require $\gamma$ and $\sigma$ to nonzero, one can show that the solutions we provided remain global minimizers in the zero limit by applying Theorem 2 from \cite{ziyin2022exactb}.} For shorthand, we use the notation $A_0:= \E[xx^T]$, and the largest and the smallest eigenvalues of $A_0$ are denoted as $a_{\rm max}$ and $a_{\rm min}$ respectively. $a_i$ denotes the $i$-th eigenvalue of $A_0$ viewed in any order. For now, it is sufficient for us to assume that the global minimum of Eq.~\eqref{eq: main loss two layer} always exists. We will prove a more general result in Proposition~\ref{prop: existence of the global minimum}, when we deal with multilayer nets.

\vspace{-1mm}
\subsection{Main Result}
\vspace{-1mm}
We first present two lemmas showing that the global minimum can only lie on a rather restrictive subspace of all possible parameter settings due to invariances in the objective. 
\begin{lemma}\label{lemma 1}
    At the global minimum of Eq.~\eqref{eq: main loss two layer}, $U_j^2 =  \frac{\gamma_w}{\gamma_u}\sum_i W_{ji}^2$ for all $j$.
\end{lemma}
\textit{Proof Sketch}. We use the fact that the first term of Eq.~\eqref{eq: main loss two layer} is invariant to a simultaneous rescaling of rows of the weight matrix to find the optimal rescaling, which implies the lemma statement. $\square$

This lemma implies that for all $j$, $|U_j|$ must be proportional to the norm of its corresponding row vector in $W$. This lemma means that using weight decay makes all layers of a deep neural network balanced. This lemma has been referred to as the ``weight balancing" condition in recent works \citep{tanaka2020pruning}, and, in some sense, is a unique and potentially essential feature of neural networks that encourages a sparse solution \citep{ziyin2022sparsity}. The following lemma further shows that, at the global minimum, all elements of $U$ must be equal. 

\begin{lemma}\label{lemma 2}
    At the global minimum, for all $i$ and $j$, we have
    \begin{equation}
        \begin{cases}
            U_i^2 = U_j^2;\\
            U_i W_{i:} = U_j W_{j:}.
        \end{cases}
    \end{equation}
\end{lemma}
\textit{Proof Sketch}. We show that if the condition is not satisfied, then an ``averaging" transformation will strictly decrease the objective. $\square$

This lemma can be seen as a formalization of the intuition suggested in the original dropout paper \citep{srivastava2014dropout}. Namely, using dropout encourages the neurons to be independent of one another and results in an averaging effect. The second lemma imposes strong conditions on the solution of the problem, and the essence of this lemma is the reduction of the original problem to a lower dimension. We are now ready to prove our first main result.

\begin{theorem}\label{theo: main theo two layer}
    The global minimum $U_*$ and $W_*$ of Eq.~\eqref{eq: main loss two layer} is $U_* = 0$ and $W_* = 0$ if and only if
    \begin{equation}
        ||\E[xy]||^2 \leq \gamma_u\gamma_w.
    \end{equation}
    When $||\E[xy]||^2 > \gamma_u\gamma_w$, the global minima are
    \begin{equation}
        \begin{cases}
            U_* = b\mathbf{r};\\
            W_* =\mathbf{r}\E[xy]^T b\left[b^2\left( \sigma^2 + d_1\right)A_0 + \gamma_w I  \right]^{-1},
        \end{cases}
        \end{equation}
        where $\mathbf{r}=(\pm 1, ...,\pm 1)$ is an arbitrary vertex of a $d_1$-dimensional hypercube, and $b$ satisfies:
        \begin{equation}\label{eq:b main}
            \bigg|\bigg|\left[b^2\left( \sigma^2 + d_1\right)A_0 + \gamma_w I  \right]^{-1} \E[xy]\bigg|\bigg|^2 =\frac{\gamma_u}{\gamma_w}.
        \end{equation}
\end{theorem}

Apparently, $b=0$ is the trivial solution that has not learned any feature due to overregularization. Henceforth, we refer to this solution (and similar solutions for deeper nets) as the ``trivial" solution. We now analyze the properties of the nontrivial solution $b^*$ when it exists.
    
The condition for the solution to become nontrivial is interesting: $||\E[xy]||^2\geq \gamma_u \gamma_w$. The term $||\E[xy]||$ can be seen as the effective strength of the signal, and $\gamma_u \gamma_w$ is the strength of regularization. This precise condition means that the learning of a two-layer can be divided into two qualitatively different regimes: an ``overregularized regime" where the global minimum is trivial, and a ``feature learning regime" where the global minimum involves actual learning. Lastly, note that our main result does not specify the exact value of $b^*$. This is because $b^*$ must satisfy the condition in Eq.~\eqref{eq:b main}, which is equivalent to a high-order polynomial in $b$ with coefficients being general functions of the eigenvalues of $A_0$, whose solutions are generally not analytical by Galois theory. One special case where an analytical formula exists for $b$ is when $A_0 =\sigma_x^2I$. See Section~\ref{app sec: exact form} for more discussion.

\subsection{Bounding the General Solution}
While the solution to $b^*$ does not admit an analytical form for a general $A_0$, one can find meaningful lower and upper bounds to $b^*$ such that we can perform an asymptotic analysis of $b^*$. At the global minimum, the following inequality holds:
\begin{align}
    ||\left[b^2\left( \sigma^2 + d_1\right)a_{\rm max}I + \gamma_w I  \right]^{-1} \E[xy]||^2 \leq ||[&b^2\left( \sigma^2 + d_1\right)A_0 + \gamma_w I  ]^{-1} \E[xy]||^2 \nonumber\\
    &\leq ||\left[b^2\left( \sigma^2 + d_1\right)a_{\rm min}I + \gamma_w I  \right]^{-1} \E[xy]||^2,
\end{align}
where $a_{\rm min}$ and $a_{\rm max}$ are the smallest and largest eigenvalue of $A_0$, respectively. The middle term is equal to $\gamma_u/\gamma_w$ by the global minimum condition in \eqref{eq:b}, and so, assuming $a_{\rm min}>0$, this inequality is equivalent to the following inequality of $b^*$:
\begin{equation}
    \frac{\sqrt{\frac{\gamma_w}{\gamma_u}}||\E[xy]|| - \gamma_w}{(\sigma^2 +d_1)a_{\rm max}} \leq b_*^2\leq \frac{\sqrt{\frac{\gamma_w}{\gamma_u}}||\E[xy]|| - \gamma_w}{(\sigma^2 +d_1)a_{\rm min}}.
\end{equation}
Namely, the general solution $b^*$ should scale similarly to the homogeneous solution in Eq.~\eqref{eq: two layer homogeneous solution} if we treat the eigenvalues of $A_0$ as constants.



\vspace{-2mm}
\section{Exact Solution for An Arbitrary-Depth  Linear Net}\label{sec: arbitrary depth net}
\vspace{-1mm}
This section extends our result to multiple layers. We first derive the analytical formula for the global minimum of a general arbitrary-depth model. We then show that the landscape for a deeper network is highly nontrivial.

\subsection{General Solution}

The loss function is
\begin{equation}\label{eq: arbitrary depth main loss}
    \E_x \E_{\epsilon^{(1)}, \epsilon^{(2)},...,\epsilon^{(D)}} \left(\sum_{i,i_1,i_2,...,i_D}^{d,d_1,d_2,...d_D} U_{i_D} \epsilon^{(D)}_{i_{D}}...\epsilon_{i_2}^{(2)} W_{i_2i_1}^{(2)}\epsilon_{i_1}^{(1)} W_{i_1i}^{(1)} x_i  - y\right)^2 + \gamma_u||U||^2 + \sum_{i=1}^D\gamma_i ||W^{(i)}||^2,
\end{equation}
where all the noises $\epsilon$ are independent, and for all $i$ and $j$, $\E[\epsilon^{(i)}_j]=1$ and $\E[(\epsilon^{(i)}_j)^2]=\sigma_i^2 + 1 > 1$. We first show that for general $D$, the global minimum exists for this objective.
\begin{proposition}\label{prop: existence of the global minimum}
    For $D\geq 1$ and strictly positive $\gamma_u,\ \gamma_1,...,\gamma_D$, the global minimum for Eq.\eqref{eq: arbitrary depth main loss} exists.
\end{proposition}

Note that the positivity of the regularization strength is crucial. If one of the $\gamma_i$ is zero, the global minimum may not exist. The following theorem is our second main result.
\begin{theorem}\label{theo: main theo multilayer}
    Any global minimum of Eq.~\eqref{eq: arbitrary depth main loss} is of the form
    \begin{equation}\label{eq: general solution}
    \begin{cases}
        U = b_u \mathbf{r}_D;\\
        W^{(i)} = b_i \mathbf{r}_{i}\mathbf{r}^T_{i-1};\\
        W^{(1)} = \mathbf{r}_1\E[xy]^T (b_u\prod_{i=2}^D b_i ) \mu \left[(b_u\prod_{i=2}^D b_i)^2 s^2\left( \sigma^2 + d_1\right)A_0 + \gamma_w I  \right]^{-1},
    \end{cases}
    \end{equation}
    where $\mu = \prod_{i=2}^D d_i$, $s^2 = \prod_{i=2}^D d_i(\sigma^2 + d_i)$, $b_u\geq 0$ and $b_i\geq 0$, and $\mathbf{r}_i=(\pm 1, ...,\pm 1)$ is an arbitrary vertex of a $d_i$-dimensional hypercube for all $i$. Furthermore, let $b_1:= \sqrt{||W_{i:}||^2 / d}$ and $b_{D+1}:= b_u$, $b_i$ satisfies
    \begin{equation}\label{eq: general b main}
        \gamma_{k+1}d_{k+1} b_{k+1}^2 = \gamma_{k}d_{k-1} b_k^2.
    \end{equation}
\end{theorem}
\textit{Proof Sketch}. We prove by induction on the depth $D$. The base case is proved in Theorem~\ref{theo: main theo two layer}. We then show that for a general depth, the objective involves optimizing subproblems, one of which is a $D-1$ layer problem that follows by the induction assumption, and the other is a two-layer problem that has been solved in Theorem~\ref{theo: main theo two layer}. Putting these two subproblems together, one obtains Eq.~\eqref{eq: general solution}. $\square$

\begin{remark}
    We deal with the technical case of having a bias term for each layer in Appendix~\ref{app sec:bias}. For example, we will show that if one has preprocessed the data such that $\E[x]=0$ and $\E[y]=0$, our main results remain precisely unchanged. 
\end{remark}

The condition in Eq.~\eqref{eq: general b main} shows that the scaling factor $b_i$ for all $i$ is not independent of one another. This automatic balancing of the norm of all layers is a consequence of the rescaling invariance of the multilayer architecture and the use of weight decay. It is well-known that this rescaling invariance also exists in a neural network with the ReLU activation, and so this balancing condition is also directly relevant for ReLU networks.


Condition~\eqref{eq: general b main} implies that all the $b_i$ can be written in terms of one of the $b_i$: 
\begin{equation}
    b_u\prod_{i=2}^{D} b_i  = c_0 \sgn\left( b_u\prod_{i=2}^{D} b_i \right)|b_2^{D}| := c_0 \sgn\left( b_u\prod_{i=2}^{D} b_i \right) b^{D}
\end{equation}
where $c_0 = \frac{(\gamma_2 d_2 d_1)^{D/2}}{\sqrt{\gamma_u\prod_{i=2}^D \gamma_i}\prod_{i=2}^D d_i \sqrt{d_1}}$ and $b\geq 0$. Consider the first layer ($i=1$), Eq~\eqref{eq: general b main} shows that the global minimum must satisfy the following equation, which is equivalent to a high-order polynomial in $b$ that does not have an analytical solution in general:
\begin{equation}\label{eq: solution condition}
    ||\E[xy]^T c_0 b^D \mu \left[c_0^2b^{2D}s^2\left( \sigma^2 + d_1\right)A_0 + \gamma_w I  \right]^{-1} ||^2 = d_2b^2.
\end{equation}
Thus, this condition is an extension of the condition~\eqref{eq:b main} for two-layer networks.

At this point, it pays to clearly define the word ``solution," especially given that it has a special meaning in this work because it now becomes highly nontrivial to differentiate between the two types of solutions.
\begin{definition}
    We say that a non-negative real $b$ is a solution if it satisfies Eq.~\eqref{eq: solution condition}. A solution is trivial if $b=0$ and nontrivial otherwise.
\end{definition}
Namely, a global minimum must be a solution, but a solution is not necessarily a global minimum. We have seen that even in the two-layer case, the global minimum can be the trivial one when the strength of the signal is too weak or when the strength of regularization is too strong. It is thus natural to expect $0$ to be the global minimum under a similar condition, and one is interested in whether the condition becomes stronger or weaker as the depth of the model is increased. However, it turns out this naive expectation is not true. In fact, when the depth of the model is larger than $2$, the condition for the trivial global minimum becomes highly nontrivial.

The following proposition shows why the problem becomes more complicated. In particular, we have seen that in the case of a two-layer net, some elementary argument has helped us show that the trivial solution $b=0$ is either a saddle or the global minimum. However, the proposition below shows that with $D\geq 2$, the landscape becomes more complicated in the sense that the trivial solution is always a local minimum, and it becomes difficult to compare the loss value of the trivial solution with the nontrivial solution because the value of $b^*$ is unknown in general.

\begin{proposition}\label{prop: 0 is local minimum}
    Let $D\geq 2$ in Eq.~\eqref{eq: arbitrary depth main loss}. Then, the solution $U=0$, $W^{(D)}=0$, ..., $W^{(1)}=0$ is a local minimum with a diagonal positive-definite Hessian $\gamma I$.
\end{proposition}

Comparing the Hessian of $D\geq 2$ and $D=1$, one notices a qualitative difference: for $D\geq 2$, the Hessian is always diagonal (at $0$); for $D=1$, in sharp contrast, the off-diagonal terms are nonzero in general, and it is these off-diagonal terms that can break the positive-definiteness of the Hessian. This offers a different perspective on why there is a qualitative difference between $D=1$ and $D=2$.

Lastly, note that, unlike the depth-$1$ case, one can no longer find a precise condition such that a $b\neq 0$ solution exists for a general $A_0$. The reason is that the condition for the existence of the solution is now a high-order polynomial with quite arbitrary intermediate terms. The following proposition gives a sufficient but stronger-than-necessary condition for the existence of a nontrivial solution, when all the $\sigma_i$, intermediate width $d_i$ and regularization strength $\gamma_i$ are the same.\footnote{This is equivalent to setting $c_0 = \sqrt{d_0}$. The result is qualitatively similar but involves additional factors of $c_0$ if $\sigma_i$,  $d_i$, and $\gamma_i$ all take different values. We thus only present the case when $\sigma_i$, $d_i$, and $\gamma_i$ are the same for notational concision and for emphasizing the most relevant terms. Also, note that this proposition gives a \textit{sufficient and necessary} condition if $A_0 = \sigma_x^2 I$ is proportional to the identity.}

\begin{proposition}\label{prop: general solution condition and property}
    Let $\sigma^2_i =\sigma^2 > 0$, $d_i=d_0$ and $\gamma_i=\gamma>0$ for all $i$. Assuming $a_{\rm min}>0$, the only solution is trivial if
    \begin{equation}\label{eq: nonexistence condition}
        \frac{D+1}{2D}||\E[xy]||d_0^{D-1}\left(\frac{(D-1)||\E[xy]||}{2D d_0 (\sigma^2 +d_0)^D a_{\rm min}} \right)^{\frac{D-1}{D+1}} < \gamma.
    \end{equation}
    Nontrivial solutions exist if
    \begin{equation}\label{eq: existence condition}
        \frac{D+1}{2D}||\E[xy]||d_0^{D-1}\left(\frac{(D-1)||\E[xy]||}{2D d_0 (\sigma^2 +d_0)^D a_{\rm max}} \right)^{\frac{D-1}{D+1}} \geq \gamma.
    \end{equation}
    Moreover, the nontrivial solutions are both lower and upper-bounded:\footnote{For $D=1$, we define the lower-bound as $\lim_{\eta\to 0^+}\lim_{D\to 1^+} \frac{1}{d_0}\left[\frac{\gamma + \eta}{||\E[xy]||}\right]^{\frac{1}{D-1}}$, which equal to zero if $\E[xy] \geq \gamma$, and $\infty$ if $\E[xy]< \gamma$. With this definition, this proposition applies to a two-layer net as well.}
    \begin{equation}\label{eq: solution upper bound}
        \frac{1}{d_0}\left[\frac{\gamma}{||\E[xy]||}\right]^{\frac{1}{D-1}} \leq b^* \leq \left[\frac{||\E[xy]||}{ d_0(\sigma^2 + d_0)^D a_{\rm max}}\right]^{\frac{1}{D+1}}.
    \end{equation}
\end{proposition}
\textit{Proof Sketch}. The proof follows from the observation that the l.h.s. of Eq.~\eqref{eq: solution condition} is a continuous function and must cross the r.h.s. under certain sufficient conditions. $\square$

One should compare the general condition here with the special condition for $D=1$. One sees that for $D\geq 2$, many other factors (such as the width, the depth, and the spectrum of the data covariance $A_0$) come into play to determine the existence of a solution apart from the signal strength $\E[xy]$ and the regularization strength $\gamma$.

\subsection{Which Solution is the Global Minimum?}
Again, we set $\gamma_i =\gamma>0$, $\sigma^2_i =\sigma^2 > 0$ and $d_i =d_0>0$ for all $i$ for notational concision. Using this condition and applying Lemma~\ref{lemma 3} to Theorem~\ref{theo: main theo multilayer}, the solution now takes the following form, where $b\geq 0$,
\begin{equation}
\label{eq: homogeneous special solution D layer b}
    \begin{cases}
        U =  \sqrt{d_0} b \mathbf{r}_{D};\\
        W^{(i)} = b \mathbf{r}_i\mathbf{r}_{i-1}^T;\\
        W^{(1)} = 
        \mathbf{r}_1\E[xy]^T d_0^{D - \frac{1}{2}} b^D \left[d_0^D ( \sigma^2 + d_0)^D b^{2D} A_0 + \gamma \right]^{-1}.
    \end{cases}
\end{equation}
The following theorem gives a sufficient condition for the global minimum to be nontrivial. It also shows that the landscape of the linear net becomes complicated and can contain more than $1$ local minimum.

\begin{theorem}\label{theo: general global minimum condition}
    Let $\sigma^2_i =\sigma^2 > 0$, $d_i=d_0$ and $\gamma_i=\gamma>0$ for all $i$ and assuming $a_{\rm min}>0$. Then, if \ifx there exists a constant $C>0$ such that for any 
    \begin{equation}
        0<\gamma <  \left[\frac{1}{Dd_0^2 C^{1/D}} \frac{1}{4}\frac{d_0^D}{(\sigma^2 + d)^D} ||\E[xy]||_{A_0^{-1}}^2 \right]^{\frac{D}{D+1}},
    \end{equation}\fi
    \begin{equation}
        ||\E[xy]||^2  \ge \frac{\gamma^\frac{D+1}{D} D^2 (\sigma^2 + d_0)^{D-1} a_{max}^\frac{D-1}{D}}{d_0^{D-1}(D-1)^\frac{D-1}{D}}
        \label{eq:condition non trivial deep}
    \end{equation}
    the global minimum of Eq.~\eqref{eq: arbitrary depth main loss} is one of the nontrivial solutions.
\end{theorem}

While there are various ways this bound can be improved, it is general enough for our purpose. In particular, one sees that, for a general depth, the condition for having a nontrivial global minimum depends not only on the $\E[xy]$ and $\gamma$ but also on the model architecture in general. For a more general architecture with different widths etc., the architectural constant $c_0$ from Eq.~\eqref{eq: solution condition} will also enter the equation. In the limit of $D\to 1^+$, relation \eqref{eq:condition non trivial deep} reduces to 
\begin{equation}
    ||\E[xy]||^2 \ge \gamma^2,
\end{equation}
which is the condition derived for the 2-layer case.

\begin{figure}[t!]
    \centering
    \vspace{-1em}
    \includegraphics[width=0.3\linewidth]{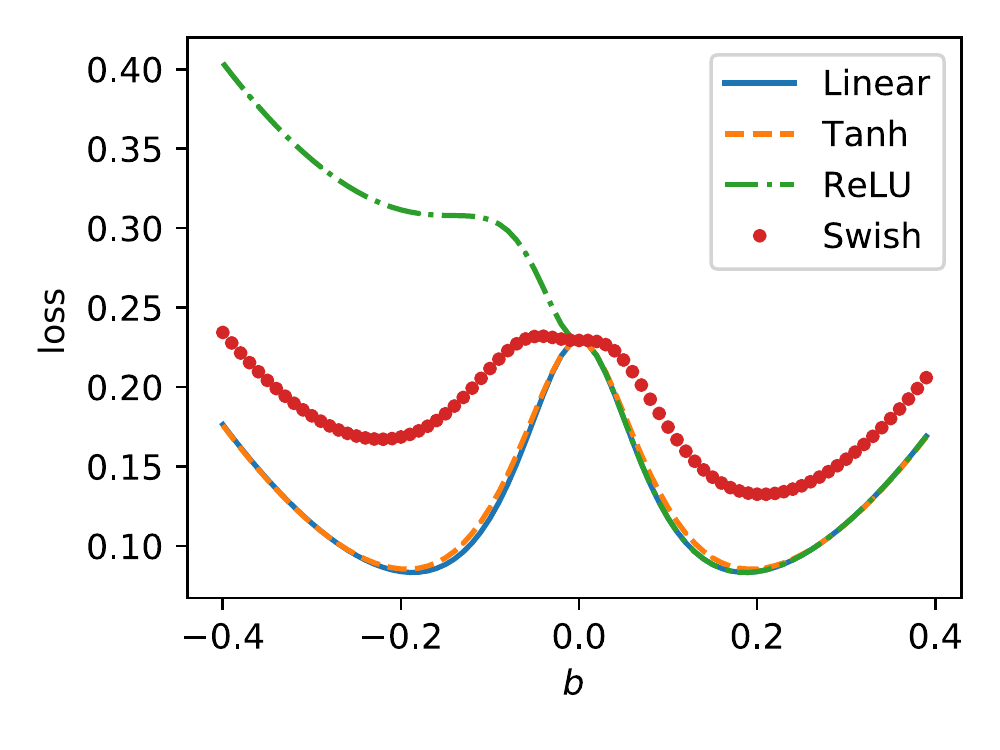}
    \includegraphics[width=0.3\linewidth]{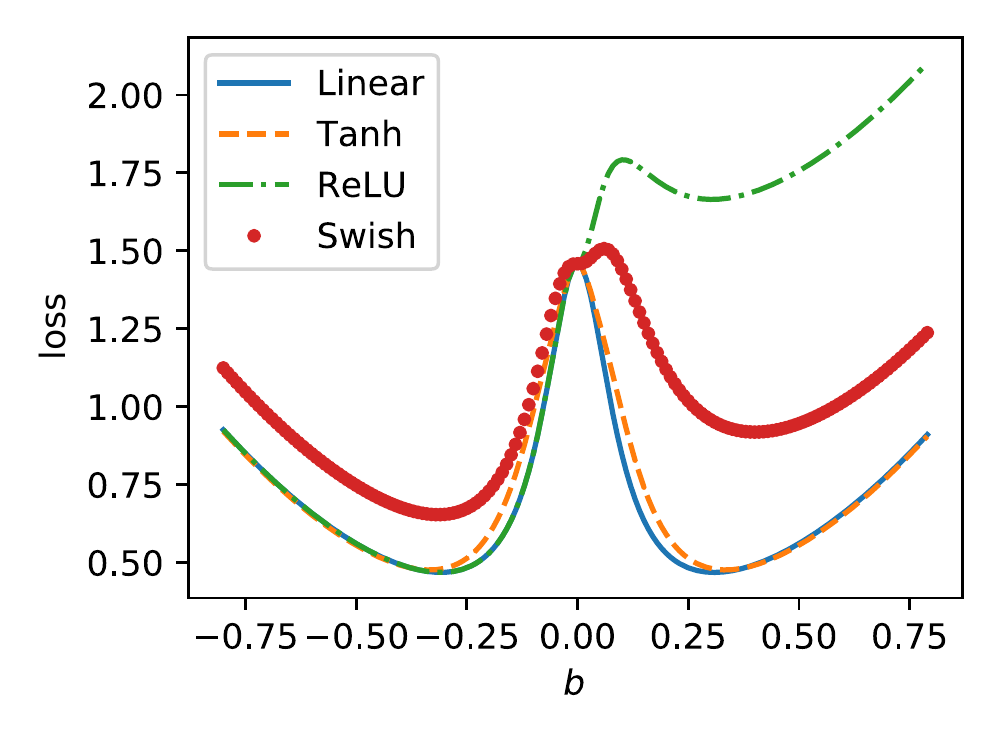}
    \vspace{-0.5em}
    \caption{\small The training loss as a function of $b$ for a $D=1$ network with different activation functions in the hidden layer. For simplicity, dropout is not implemented. The non-linear activation functions we considered are ReLU, Tanh, and Swish. The left and right panels use different data. \textbf{Left:} $X$ are Gaussian random vectors, and $y = v \cdot x$ is a linear function of $x$. \textbf{Right:} $x$ are Gaussian random vectors, and $y = v \cdot tanh(x)$ are nonlinear functions of data; the weight $v$ is obtained as a Gaussian random vector. }
    \label{fig: non linear}
    \vspace{-1em}
\end{figure}

\vspace{-2mm}
\section{Implications}\label{sec: implications}
\vspace{-1mm}
\textbf{Relevance to nonlinear models.} {We first caution the readers that the following discussion should be taken with a caveat and is based on the philosophy that deep linear nets can approximate the nonlinear ones. This approximation certainly holds for fully connected models with differentiable activation functions such as tanh or Swish because they are, up to first-order Taylor expansion, a deep linear net around zero, which is the region for which our theory is the most relevant. We empirically demonstrate that close to the origin, the landscape of linear nets can indeed approximate that of nonlinear nets quite well.} To compare, we plug in the solution in Theorem \ref{prop: main theo multilayer} to both linear and nonlinear models of the same architecture and compare the loss values at different values of $b$ around $b=0$. For simplicity, we only consider the case $D=1$. The activation functions we consider are ReLU, Tanh, and Swish \citep{ramachandran2017searching}, a modern and differentiable variant of ReLU. See  Fig.~\ref{fig: non linear}.

The regressor $x\in\mathbb{R}^d$ is sampled as Gaussian random vectors. We consider two methods of generating $y$; the first one (left) is $y = v \cdot x$. The second one (right) is $y = v \cdot \tanh(x)$, where the weight $v\in\mathbb{R}^d$ is obtained as a Gaussian random vector. Fig.~\ref{fig: non linear} shows that the landscape consisting of Tanh is always close to the linear landscape. Swish is not as good as Tanh, but the Swish landscape shows a similar tendency to the linear landscape. The ReLU landscape is not so close to the linear landscape either for $b > 0$ or $b < 0$, but it agrees completely with the linear landscape on the other side, as expected. Besides the quantitative closeness, it is also important to note that all the landscapes agree qualitatively, containing the same number of local minima at similar values of $b$.

\textbf{Landscape of multi-layer neural networks}. The combination of Theorem~\ref{theo: general global minimum condition} and Proposition~\ref{prop: 0 is local minimum} shows that the landscape of a deep neural network can become highly nontrivial when there is a weight decay and when the depth of the model is larger than $2$. This gives an incomplete but meaningful picture of a network's complicated but interesting landscape beyond two layers (see Figure~\ref{fig:summary} for an incomplete summary of our results). In particular, even when the nontrivial solution is the global minimum, the trivial solution is still a local minimum that needs to be escaped. Our result suggests the previous understanding that all local minima of a deep linear net are global cannot generalize to many practical settings where deep learning is found to work well. For example, a series of works attribute the existence of bad (non-global) minima to the use of nonlinearities \citep{kawaguchi2016deep} or the use of a non-regular (non-differentiable) loss function \citep{laurent2018deep}. Our result, in contrast, shows that the use of a simple weight decay is sufficient to create a bad minimum.\footnote{Some previous works do suggest the existence of bad minima when weight decay is present, but no direct proof exists yet. For example, \cite{taghvaei2017regularization} shows that when the model is approximated by a linear dynamical system, regularization can cause bad local minima. \cite{mehta2021loss} shows the existence of bad local minima in deep linear networks with weight decay through numerical simulations.} Moreover, the problem with such a minimum is two-fold: (1) (optimization) it is not global and so needs to be ``overcome" and (2) (generalization) it is a minimum that has not learned any feature at all because the model constantly outputs zero. To the best of our knowledge, previous to our work, there has not been any proof that a bad minimum can generically exist in a rather arbitrary network without any restriction on the data.\footnote{In the case of nonlinear networks without regularization, a few works proved the existence of bad minima. However, the previous results strongly depend on the data and are rather independent of architecture. For example, one major assumption is that the data cannot be perfected and fitted by a linear model \citep{yun2018small, liu2021spurious, he2020piecewise}. Some other works explicitly construct data distribution \citep{safran2018spurious, venturi2019spurious}. Our result, in contrast, is independent of the data.} Thus, our result offers direct theoretical justification for the widely believed importance of escaping local minima in the field of deep learning \citep{kleinberg2018alternative, liu2021noise, mori2022power}. In particular, previous works on escaping local minima often hypothesize landscapes that are of unknown relevance to an actual neural network. With our result, this line of research can now be established with respect to landscapes that are actually deep-learning-relevant.

Previous works also argue that having a deeper depth does not create a bad minimum \citep{lu2017depth}. While this remains true, its generality and applicability to practical settings now also seem low. Our result shows that as long as weight decay is used, and as long as $D\geq 2$, there is indeed a bad local minimum at $0$. In contrast, there is no bad minimum at $0$ for a depth-$2$ network: the point $b=0$ is either a saddle or the global minimum.\footnote{Of course, in practice, the model trained with SGD can still converge to the trivial solution even if it is a saddle point \citep{ziyin2021sgd} because minibatch SGD is, in general, not a good estimator of the local minima.} Having a deeper depth thus alters the qualitative nature of the landscape, and our results agree better with the common observation that a deeper network is harder, if not impossible, to optimize.

{\color{black}We note that our result can also be relevant for more modern architectures such as the ResNet. Using ResNet, one needs to change the dimension of the hidden layer after every bottleneck, and a learnable linear transformation is applied here. Thus, the “effective depth” of a ResNet would be roughly between the number of its bottlenecks and its total number of blocks. For example, a ResNet18 applied to CIFAR10 often has five bottlenecks and 18 layers in total. We thus expect it to have qualitatively similar behavior to a deep linear net with a depth in between. See Figure~\ref{fig:summary}. The experimental details are given in Section~\ref{app sec; experimental details}.}

\textbf{Learnability of a neural network}. Now we analyze the solution when $D$ tends to infinity. We first note that the existence condition bound in \eqref{eq: existence condition} becomes exponentially harder to satisfy as $D$ becomes large:
\begin{equation}\label{eq: asymptotic existence condition}
    ||\E[xy]||^2 \geq 4 d_0^2 a_{\rm max}\gamma  e^ {D \log[(\sigma^2 + d_0)/d_0]}   + O(1)  .
\end{equation}
When this bound is not satisfied, the given neural network cannot learn the data. Recall that for a two-layer net, the existence condition is nothing but $||\E[xy]||^2>\gamma^2$, independent of the depth, width, or stochasticity in the model. For a deeper network, however, every factor comes into play, and the architecture of the model has a strong (and dominant) influence on the condition. In particular, a factor that increases polynomially in the model width and exponentially in the model depth appears. 

A practical implication is that the use of weight decay may be too strong for deep networks. If one increases the depth or width of the model, one should also roughly decrease $\gamma$ according to Eq.~\eqref{eq: asymptotic existence condition}.

\begin{wrapfigure}{r}{0.38\textwidth}  \begin{center}
\vspace{-1.5em}
    \includegraphics[width=\linewidth]{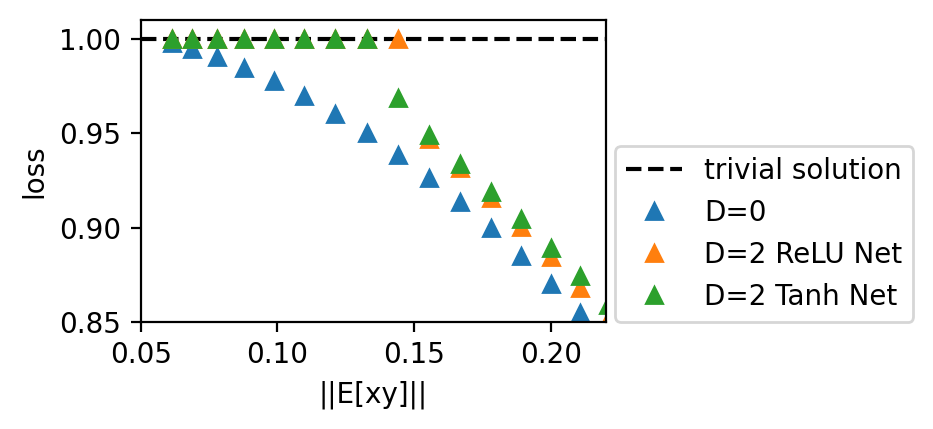}
    \vspace{-1.5em}
    \caption{\small Training loss of $D=2$ neural networks with ReLU and Tanh activations across synthetic tasks with different $||\E[xy]||$. We see that with the Kaiming initialization, both the Tanh net and the ReLU net are stuck at the trivial solution in expectation of our theory. In contrast, an optimized linear regressor ($D=0$) is better than the trivial solution when $||\E[xy]||>0$. See Section~\ref{app sec; experimental details} for experimental details.}\label{fig: learnability}
  \end{center}
  \vspace{-1em}
\end{wrapfigure}

\textbf{Insufficiency of the existing initialization schemes}. We have shown that $0$ is often a bad local minimum for deep learning. Our result further implies that escaping this local minimum can be highly practically relevant because standard initialization schemes are trapped in this local minimum for tasks where the signal $\E[xy]$ is weak. See Inequality~\eqref{eq: solution upper bound}: any nontrivial global minimum is lower-bounded by a factor proportional to $(\gamma/||\mathbb{E}[xy]||^{1/(D-1)})/ d_0 $, which can be seen as an approximation of the radius of the local minimum at the origin. In comparison, standard deep learning initialization schemes such as Kaiming init. initialize at a radius roughly $1/\sqrt{d_0}$. Thus, for tasks $\mathbb{E}[xy]\ll \gamma/\sqrt{d_0}$, these initialization methods are likely to initialize the model in the basin of attraction of the trivial regime, which can cause a serious failure in learning. To demonstrate, we perform a numerical simulation shown in the right panel of Figure~\ref{fig: learnability}, where we train  $D=2$ nonlinear networks with width $32$ with SGD on tasks with varying $||\mathbb{E}[xy]||$. For sufficiently small $||\mathbb{E}[xy]||$, the model clearly is stuck at the origin.\footnote{There are many natural problems where the signal is extremely weak. One well-known example is the problem of future price prediction in finance, where the fundamental theorem of finance forbids a large $||\E[xy]||$ \citep{fama1970efficient}.} In contrast, linear regression is never stuck at the origin. Our result thus suggests that it may be desirable to devise initialization methods that are functions of the data distribution.

\textbf{Prediction variance of stochastic nets}. A major extension of the standard neural networks is to make them stochastic, namely, to make the output a random function of the input. In a broad sense, stochastic neural networks include neural networks trained with dropout \citep{srivastava2014dropout, gal2016dropout}, Bayesian networks \citep{mackay1992bayesian}, variational autoencoders (VAE) \citep{kingma2013auto}, and generative adversarial networks \citep{goodfellow2014generative}.  Stochastic networks are thus of both practical and theoretical importance to study. Our result can also be used for studying the theoretical properties of stochastic neural networks. Here, we present a simple application of our general solution to analyze the properties of a stochastic net. The following theorem summarizes our technical results.

\begin{theorem}\label{theo: variance}
    Let $\sigma^2_i =\sigma^2 > 0$, $d_i=d_0$ and $\gamma_i=\gamma>0$ for all $i$. Let $A_0=\sigma_x^2I$. Then, at any global minimum of Eq.~\eqref{eq: arbitrary depth main loss}, holding other parameters fixed,
    \begin{enumerate}[noitemsep,topsep=0pt,parsep=0pt,partopsep=0pt]
        \item in the limit of large $d_0$, $\V[f(x)]  = O\left(d_0^{-1}\right);$
        \item in the limit of large $\sigma^2$, $\V[f(x)] = O \left( \frac{1}{(\sigma^2)^{D}} \right)$;
        \item In the limit of large $D$, $\V[f(x)] = O \left( e^{-2D\log[(\sigma^2 + d_0)/d_0]} \right)$.
    \end{enumerate}
\end{theorem}

Interestingly, the scaling of prediction variance in asymptotic $\sigma^2$ is different for different widths. The third result shows that the prediction variance decreases exponentially fast in $D$. In particular, this result answers a question recently proposed in \cite{ziyin2022stochastic}: does a stochastic net trained on MSE have a prediction variance that scales towards $0$? 
We improve on their result in the case of a deep linear net by (a) showing that the $d_0^{-1}$ is tight in general, independent of the depth or other factors of the model, and (b) proving a bound showing that the variance also scales towards zero as depth increases, which is a novel result of our work. Our result also offers an important insight into the cause of the vanishing prediction variance. Previous works \citep{alemi2018fixing} often attribute the cause to the fact that a wide neural network is too expressive. However, our result implies that this is not always the case because a linear network with limited expressivity can also have a vanishing variance as the model tends to an infinite width.

\textbf{Collapses in deep learning.} Lastly, we comment briefly on the apparent similarity between different types of collapses that occur in deep learning. For neural collapse, our result agrees with the recent works that identify weight decay as a main cause \citep{rangamani2022neural}. For Bayesian deep learning, \cite{wang2022posterior} identified the cause of the posterior collapse in a two-layer VAE structure to be that the regularization of the mean of the latent variable $z$ is too strong. More recently, the origin and its stability have also been discussed as the dimensional collapse in self-supervised learning \citep{ziyin2023what}. Although appearing in different contexts of deep learning, the three types of collapses share the same phenomenology that the model converges to a ``collapsed" regime where the learned representation becomes low-rank or constant, which agrees with the behavior of the trivial regime we identified. We refer the readers to \cite{ziyin2022exactb} for a study of how the second-order phase transition framework of statistical physics can offer a possible unified explanation of these phenomena.

\vspace{-2mm}
\section{Conclusion}
\vspace{-1mm}
In this work, we derived the exact solution of a deep linear net with arbitrary depth and width and with stochasticity. Our work sheds light on the highly complicated landscape of a deep neural network. Compared to the previous works that mostly focus on the qualitative understanding of the linear net, our result offers a more precise quantitative understanding of deep linear nets. Quantitative understanding is one major benefit of knowing the exact solution, whose usefulness we have also demonstrated with the various implications. The results, although derived for linear models, are also empirically shown to be relevant for networks with nonlinear activations. Lastly, our results strengthen the line of thought that analytical approaches to deep linear models can be used to understand deep neural networks, and it is the sincere hope of the authors to attract more attention to this promising field.


\section*{Acknowledgement}
Ziyin is financially supported by the GSS scholarship of the University of Tokyo and the JSPS fellowship. Li is financially supported by CNRS. X. Meng is supported by JST CREST Grant Number JPMJCR1912, Japan. 

\bibliographystyle{apalike}
\bibliography{ref}

\section*{Checklist}

\begin{enumerate}

\item For all authors...
\begin{enumerate}
  \item Do the main claims made in the abstract and introduction accurately reflect the paper's contributions and scope?
    \answerYes{}
  \item Did you describe the limitations of your work?
    \answerYes{}
  \item Did you discuss any potential negative societal impacts of your work?
    \answerNA{}
  \item Have you read the ethics review guidelines and ensured that your paper conforms to them?
    \answerYes{}
\end{enumerate}

\item If you are including theoretical results...
\begin{enumerate}
  \item Did you state the full set of assumptions of all theoretical results?
    \answerYes{}
        \item Did you include complete proofs of all theoretical results?
    \answerYes{See Appendix.}
\end{enumerate}

\item If you ran experiments...
\begin{enumerate}
  \item Did you include the code, data, and instructions needed to reproduce the main experimental results (either in the supplemental material or as a URL)?
    \answerNo{The experiments are only for demonstration and are straightforward to reproduce following the theory.}
  \item Did you specify all the training details (e.g., data splits, hyperparameters, how they were chosen)?
    \answerYes{}
        \item Did you report error bars (e.g., with respect to the random seed after running experiments multiple times)?
    \answerNo{The fluctuations are visually negligible.}
        \item Did you include the total amount of compute and the type of resources used (e.g., type of GPUs, internal cluster, or cloud provider)?
    \answerYes{They are done on a single 3080Ti GPU.}
\end{enumerate}

\item If you are using existing assets (e.g., code, data, models) or curating/releasing new assets...
\begin{enumerate}
  \item If your work uses existing assets, did you cite the creators?
    \answerNA{}
  \item Did you mention the license of the assets?
    \answerNA{}
  \item Did you include any new assets either in the supplemental material or as a URL?
    \answerNA{}
  \item Did you discuss whether and how consent was obtained from people whose data you're using/curating?
    \answerNA{}
  \item Did you discuss whether the data you are using/curating contains personally identifiable information or offensive content?
    \answerNA{}
\end{enumerate}

\item If you used crowdsourcing or conducted research with human subjects...
\begin{enumerate}
  \item Did you include the full text of instructions given to participants and screenshots, if applicable?
    \answerNA{}
  \item Did you describe any potential participant risks, with links to Institutional Review Board (IRB) approvals, if applicable?
    \answerNA{}
  \item Did you include the estimated hourly wage paid to participants and the total amount spent on participant compensation?
    \answerNA{}
\end{enumerate}

\end{enumerate}

\clearpage
\appendix

\section{Experimental Details}\label{app sec; experimental details}
For the experiment in Figure~\ref{fig: learnability}, the input data consists of $1000$ data points sampled from a multivariate Gaussian distribution: $x\sim \mathcal{N}(0, I_5)$. The target is generated by a linear transformation $y= v\cdot x$, where the norm of $v$ is rescaled to obtain different values of $||\E[xy]||$ as the control parameter of the simulation. The models are with $D=2$ neural networks with bias terms and with hidden width $32$ for both hidden layers. The training proceeds with gradient descent with a learning rate of $0.1$ for $10^4$ iterations when the training loss has stopped decreasing for all the experiments.

For the CIFAR10 experiments, we train a standard ResNet18 with roughly $10^7$ parameters under the standard procedure, with a batch size of $256$ for $100$ epochs.\footnote{Specifically, we use the implementation and training procedure of \url{https://github.com/kuangliu/pytorch-cifar}, with standard augmentations such as random crop, etc.} For the linear models, we use a hidden width of $32$ without any bias term. The training proceeds with SGD with batch size $256$ for $100$ epochs with a momentum of $0.9$. The learning rate is $0.002$, chosen as the best learning rate from a grid search over $[0.001, 0.002,..., 0.01]$.

\section{Proofs}\label{app sec: proof}
\subsection{Proof of Lemma~\ref{lemma 1}}\label{app sec: lemma 1 proof}
\textit{Proof}. Note that the first term in the loss function is invariant to the following rescaling for any $a> 0$: 
\begin{equation}
    \begin{cases}
        U_i\to aU_i;\\
        W_{ij} \to W_{ij}/a;
    \end{cases}
\end{equation}
meanwhile, the $L_2$ regularization term changes as $a$ changes. Therefore, the global minimum must have a minimized $a$ with respect to any $U$ and $W$.

One can easily find the solution:
\begin{equation}
    a^* = \arg\min_a \left(\gamma_u a^2 U_i^2 + \gamma_w \sum_j \frac{W_{ij}^2}{a^2}\right) = \left(\frac{\gamma_w \sum_j W_{ij}^2}{\gamma_u U_i^2}\right)^{1/4}.
\end{equation}
Therefore, at the global minimum, we must have $\gamma_u a^2 U_i^2 = \gamma_w \sum_j \frac{W_{ij}^2}{a^2}$, so that 
\begin{equation}
    (U_i^*)^2 = (a^*U_i)^2 = \frac{\gamma_w}{\gamma_u} \sum_j (W_{ij}^*)^2,
\end{equation}
which completes the proof. $\square$

\subsection{Proof of Lemma~\ref{lemma 2}}\label{app sec: lemma 2 proof}
\textit{Proof}. By Lemma~\ref{lemma 1}, we can write $U_i$ as $b_i$ and $W_{i:}$ as $b_i w_i$ where $w_i$ is a unit vector, and finding the global minimizer of Eq.~\eqref{eq: main loss two layer} is equivalent to finding the minimizer of the following objective,
\begin{align}
&\mathbb{E}_{x, \varepsilon} \left[ \left(\sum_{i,j} b^2_{i} \epsilon_i w_{ij}x_j - y \right)^2\right] +  (\gamma_u + \gamma_w)||b||_2^2,\\
&= \mathbb{E}_{x} \left[ \left(\sum_{i,j} b^2_{i} w_{ij}x_j - y \right)^2\right] + \sigma^2\sum_{ij} b_i^4 \left(\sum_{k} w_{ik}x_k\right)^2 +   (\gamma_u + \gamma_w)||b||_2^2,\label{eq: equivalent loss in lemma 2}
\end{align}
The lemma statement is equivalent to $b_i=b_j$ for all $i$ and $j$.

We prove this by contradiction. Suppose there exist $i$ and $j$ such that $b_i\neq b_j$, we can choose $i$ to be the index of $b_i$ with maximum $b_i^2$, and let $j$ be the index of $b_j$ with minimum $b_j^2$. Now, we can construct a different solution by the following replacement of  $b_iw_{i:}$ and $b_jw_{j:}$:
\begin{equation}
    \begin{cases}
        b_i^2w_{i:}\to c^2 v;\\
        b_j^2w_{j:}\to c^2 v,
    \end{cases}
\end{equation}
where $c$ is a positive scalar and $v$ is a unit vector such that $2c^2 v = b_i^2w_{i:} + b_j^2w_{j:}$. Note that, by the triangular inequality, $2c^2 \leq b_i^2 + b_j^2$. Meanwhile, all the other terms, $b_k$ for $k\neq i$ and $k\neq j$, are left unchanged. This transformation leaves the first term in the loss function \eqref{eq: equivalent loss in lemma 2} unchanged, and we now show that it decreases the other terms.

The change in the second term is
\begin{equation}
    \left(b_i^2 \sum_k w_{ik}x_k \right)^2 + \left(b_j^2 \sum_k w_{jk}x_k \right)^2 \to  2\left(c^2 \sum_k v_{k}x_k \right)^2 = \frac{1}{2}\left(b_i^2\sum_k w_{ik}x_k + b_j^2\sum_{k}w_{jk}x_k \right)^2.
\end{equation}
By the inequality $a^2 + b^2 \geq (a+b)^2/2$, we see that the left-hand side is larger than the right-hand side.

We now consider the $L_2$ regularization term. The change is
\begin{equation}
    (\gamma_u + \gamma_w)(b_i^2 + b_j^2) \to 2(\gamma_u + \gamma_w)c^2,
\end{equation}
and the left-hand side is again larger than the right-hand side by the inequality mentioned above: $2c^2 \leq b_i^2 + b_j^2$. Therefore, we have constructed a solution whose loss is strictly smaller than that of the global minimum: a contradiction. Thus, the global minimum must satisfy
\begin{equation}
    U_i^2=U_j^2
\end{equation}
for all $i$ and $j$. 

Likewise, we can show that $U_iW_{i:} = U_jW_{j:}$ for all $i$ and $j$. This is because the triangular inequality $2c^2 \leq b_i^2 + b_j^2$ is only an equality if $U_iW_{i:} = U_jW_{j:}$. If $U_i W_{i:} \neq U_j W_{j:}$, following the same argument above, we arrive at another contradiction. 
$\square$

\subsection{Proof of Theorem~\ref{theo: main theo two layer}}
\textit{Proof}. By Lemma~\ref{lemma 2}, at any global minimum, we can write $U_* = b\mathbf{r}$ for some $b\in \mathbb{R}$. We can also write $W_* =\mathbf{r}v^T$ for a general vector $v\in \mathbb{R}^{d}$. Without loss of generality, we assume that $b>0$ (because the sign of $b$ can be absorbed into $\mathbf{r}$).

The original problem in Eq.~\eqref{eq: main loss two layer} is now equivalently reduced following problem because $\mathbf{r}^T \mathbf{r} = d_1$:
\begin{align}
\min_{b,v}\mathbb{E}_{x} \left[ \left(bd_1 \sum_{j}v_jx_j - y \right)^2 +  b^2d_1 \sigma^2\left(\sum_{k} v_{k}x_k\right)^2\right] +   \gamma_u d_1 b^2 + \gamma_w d_1 ||v||_2^2.
\end{align}

For any fixed $b$, the global minimum of $v$ is well known:\footnote{Namely, it is the solution of a ridge linear regression problem.}
\begin{equation}
    v = b \E[xy]^T\left[b^2\left( \sigma^2 + d_1\right)A_0 + \gamma_w I  \right]^{-1}.
\end{equation}
By Lemma~\ref{lemma 1}, at a global minimum, $b$ also satisfies the following condition:
\begin{equation}
    b^2 = \frac{\gamma_w}{\gamma_u}||v||^2,
\end{equation}
One solution to this equation is $b=0$, and we are interested in whether solutions with $b\neq 0$ exist. If there is no other solution, then $b=0$ must be the unique global minimum; otherwise, we need to identify which of the solutions are actual global minima. When $b\neq 0$,
\begin{equation}
\label{eq:b}
    \bigg|\bigg|\left[b^2\left( \sigma^2 + d_1\right)A_0 + \gamma_w I  \right]^{-1} \E[xy]\bigg|\bigg|^2 =\frac{\gamma_u}{\gamma_w}.
\end{equation}
Note that the left-hand side is monotonically decreasing in $b^2$, and is equal to $\gamma^{-2}_w ||\E[xy]||^2$ when $b=0$. When $b\to \infty$, the left-hand side tends to $0$. Because the left-hand side is a continuous and monotonic function of $b$, a unique solution $b_* > 0 $ that satisfies Eq.~\eqref{eq:b} exists if and only if $\gamma^{-2}_w ||\E[xy]||^2 >\gamma_u/\gamma_w$, or,
\begin{equation}\label{eq: two layer existence condition}
    ||\E[xy]||^2 > \gamma_u \gamma_w.
\end{equation}
Therefore, at most, three candidates for global minima of the loss function exist:
\begin{equation}
    \begin{cases}
        b = 0, \ v=0 & \text{if } ||\E[xy]||^2 \le \gamma_u \gamma_w;\\
        b= \pm b_*,\ v=b\left[b^2\left( \sigma^2 + d_1\right)A_0 + \gamma_w I  \right]^{-1} \E[xy], & \text{if }||\E[xy]||^2 > \gamma_u \gamma_w,
    \end{cases}
\end{equation}
where $b^*>0$.

In the second case, one needs to discern the saddle points from the global minima. Using the expression of $v$, one finds the expression of the loss function as a function of $b$
\begin{align}
    d_1(d_1+\sigma^2)b^4\sum_i\frac{\E[x'y]_i^2 a_{i}}{[b^2(\sigma^2+d_1)a_i + \gamma_w]^2} &- 2 b^2d_1\sum_i\frac{\E[x'y]_i^2}{b^2(\sigma^2+d_1)a_i + \gamma_w} + \E[y^2] \nonumber\\ &+ \gamma_u d_1 b^2 + \gamma_w d_1 \sum_i\frac{\E[x'y]_i^2 b^2 }{[b^2(\sigma^2+d_1)a_i + \gamma_w]^2},
\end{align}
where $x'= Rx$ such that $RA_0R^{-1}$ is a diagonal matrix. We now show that condition~\eqref{eq: two layer existence condition} is sufficient to guarantee that $0$ is not the global minimum.

At $b=0$, the first nonvanishing derivative of $b$ is the second-order derivative. The second order derivative at $b = 0$ is 
\begin{equation}
    -2 d_1 ||\E[xy]||^2 / \gamma_w + 2\gamma_u d_1,
\end{equation}
which is negative if and only if $||\E[xy]||^2 > \gamma_u \gamma_w$. If the second derivative at $b = 0$ is negative, $b=0$ cannot be a minimum. It then follows that for $||\E[xy]||^2 > \gamma_u \gamma_w$, $b= \pm b^*$, $v=b\left[b^2\left( \sigma^2 + d_1\right)A_0 + \gamma_w I  \right]^{-1} \E[xy]$ are the two global minimum (because the loss is invariant to the sign flip of $b$). For the same reason, when $||\E[xy]||^2 < \gamma_u \gamma_w$, $b=0$ gives the unique global minimum. This finishes the proof. $\square$

\subsection{Proof of Proposition~\ref{prop: existence of the global minimum}}
\textit{Proof}. We first show that there exists a constant $r$ such that the global minimum must be confined within a (closed) $r$-Ball around the origin. The objective \eqref{eq: arbitrary depth main loss} can be upper-bounded by
\begin{equation}
    Eq.~\eqref{eq: arbitrary depth main loss} \geq \gamma_u||U||^2 +\sum_{i=1}^D\gamma_i ||W^{(i)}||^2 \geq \gamma_{\rm min} \left(||U||^2 + \sum_{i}||W^{(i)}||^2\right),
\end{equation}
where $\gamma_{\rm min}:=\min_{i\in\{u, 1, 2,...,D\}} >0$. Now, let $w$ denote be the union of all the parameters ($U, W^{(i)}$) and viewed as a vector. We see that the above inequality is equivalent to
\begin{equation}
    Eq.~\eqref{eq: arbitrary depth main loss} \geq \gamma_{\rm min} ||w||^2.
\end{equation}

Now, note that the loss value at the origin is $\E[y^2]$, which means that for any $w$, whose norm $||w||^2 \geq \E[y^2]/\gamma_{\rm min}$, the loss value must be larger than the loss value of the origin. Therefore, let $r = \E[y^2]/\gamma_{\rm min}$, we have proved that the global minimum must lie in a closed $r$-Ball around the origin. 

As the last step, because the objective is a continuous function of $w$ and the $r$-Ball is a compact set, the minimum of the objective in this $r$-Ball is achievable. This completes the proof. $\square$

\subsection{Proof of Theorem~\ref{theo: main theo multilayer}}
We divide the proof into the proof of a proposition and a lemma,  and combining the following proposition and lemma obtains the theorem statement.
\subsubsection{Proposition~\ref{prop: main theo multilayer}}
\begin{proposition}\label{prop: main theo multilayer}
    Any global minimum of Eq.~\eqref{eq: arbitrary depth main loss} is of the form
    \begin{equation}
    \begin{cases}
        U = b_u \mathbf{r}_D;\\
        W^{(i)} = b_i \mathbf{r}_{i}\mathbf{r}^T_{i-1};\\
        W^{(1)} = \mathbf{r}_1\E[xy]^T (b_u\prod_{i=2}^D b_i ) \mu \left[(b_u\prod_{i=2}^D b_i)^2 s^2\left( \sigma^2 + d_1\right)A_0 + \gamma_w I  \right]^{-1},
    \end{cases}
    \end{equation}
    where $\mu = \prod_{i=2}^D d_i$, $s^2 = \prod_{i=2}^D d_i(\sigma^2 + d_i)$, $b_u\geq 0$ and $b_i\geq 0$, and $\mathbf{r}_i=(\pm 1, ...,\pm 1)$ is an arbitrary vertex of a $d_i$-dimensional hypercube for all $i$.
\end{proposition}

\textit{Proof}. Note that the trivial solution is also a special case of this solution with $b=0$. We thus focus on deriving the form of the nontrivial solution.

We prove by induction on $D$. The base case with depth $1$ is proved in Theorem~\ref{theo: main theo two layer}. We now assume that the same holds for depth $D-1$ and prove that it also holds for depth $D$.

For any fixed $W^{(1)}$, the loss function can be equivalently written as
\begin{align}
    \E_{\tilde{x}}  \E_{ \epsilon^{(2)},...,\epsilon^{(D)}} \left(\sum_{i_1,i_2,...,i_D}^{d_1,d_2,...d_D} U_{i_D} \epsilon^{(D)}_{i_{D}}...\epsilon_{i_2}^{(2)} W_{i_2i_1}^{(2)} \tilde{x}_{i_1}  - y\right)^2 + \gamma_u||U||^2 +\sum_{i=2}^D\gamma_i ||W^{(i)}||^2 + const.,
\end{align}
where $\tilde{x} = \epsilon_{i_1}^{(1)}\sum_iW^{(1)}_{i_1i}x_i$. Namely, we have reduced the problem to a problem involving only a depth $D-1$ linear net with a transformed input $\tilde{x}$.

By the induction assumption, the global minimum of this problem takes the form of Eq.~\eqref{eq: general solution}, which means that the loss function can be written in the following form:
\begin{equation}
    \E_{\tilde{x}}  \E_{ \epsilon^{(2)},...,\epsilon^{(D)}} \left( b_u b_D ...b_3\sum_{i_1,i_2,...,i_D}^{d_1,d_2,...d_D}  \epsilon^{(D)}_{i_{D}}...\epsilon_{i_2}^{(2)} v_{i_1} \tilde{x}_{i_1}  - y\right)^2 + L_2\ reg.,
\end{equation}
for an arbitrary optimizable vector $v_{i_1}$. The term $\sum_{i_2,...,i_D}^{d_2,...d_D}  \epsilon^{(D)}_{i_{D}}...\epsilon_{i_2}^{(2)}:= \eta$ can now be regarded as a single random variable such that $\E[\eta] = \prod_{i=2}^D d_i : = \mu$ and $\E[\eta^2] = \prod_{i=2}^Dd_i(\sigma_i^2 + d_i) := s^2$. Computing the expectation over all the noises except for $\epsilon^{(1)}$, one finds
\begin{align}
    &\E_{\tilde{x}}   \left( b_u b_D ...b_3 s \sum_{i_1}v_{i_1} \tilde{x}_{i_1}  - \frac{\mu y}{s}\right)^2 + L_2\ reg.+ const.\\
    &=\E_{x,\epsilon^{(1)}}  \left( b_u b_D ...b_3 s \sum_{i,i_1}v_{i_1} \epsilon_{i_1}^{(1)}W^{(1)}_{i_1i}{x}_{i}  - \frac{\mu y}{s}\right)^2 + L_2\ reg.+  const.,
\end{align}
where we have ignored the constant term because it does not affect the minimizer of the loss. Namely, we have reduced the original problem to a two-layer linear net problem where the label becomes effectively rescaled for a deep network.

For any fixed $b_u,...,b_3$, we can define $\bar{x}:= b_u b_D ...b_3 s x$, and obtain the following problem, whose global minimum we have already derived:
\begin{equation}
    \E_{\bar{x}}  \E_{ \epsilon^2,...,\epsilon_{D}} \left( \sum_{i,i_1}v_{i_1} W^{(1)}_{i_1i}\bar{x}_{i}  - \frac{\mu y}{s}\right)^2.
\end{equation}
By Theorem~\ref{theo: main theo two layer}, the global minimum is identically $0$ if $||\E[\mu\bar{x}y/s]||^2 < d_2\gamma_{2}\gamma_{1}$, or, $\E[xy] \le \frac{\gamma_{2}\gamma_{1}}{b_3^2...b_u^2 \left(\prod_{i=3}^D d_i\right)}$. When $\E[xy] > \frac{\gamma_{2}\gamma_{1}}{b_3^2...b_u^2 \left(\prod_{i=3}^D d_i\right)}$, the solution can be non-trivial:
\begin{equation}
    \begin{cases}
            v_* = b_2^*\mathbf{r}_1;\\
            W_* =\mathbf{r}_1\E[xy]^T \mu b_2^* b_3...b_u\left[(b_2^*)^2 d_3^2...d_D^2b_u^2 s^2\left( \sigma^2 + d_1\right)A_0 + \gamma_1 I  \right]^{-1},
        \end{cases}
    \end{equation}
for some $b^*_2$. This proves the theorem. $\square$

\subsection{Lemma~\ref{lemma 3}}

\begin{lemma}\label{lemma 3}
    At any global minimum of Eq.~\eqref{eq: arbitrary depth main loss},  let $b_1:= \sqrt{||W_{i:}||^2 / d}$ and $b_{D+1}:= b_u$,
    \begin{equation}
    \gamma_{k+1}d_{k+1} b_{k+1}^2 = \gamma_{k}d_{k-1} b_k^2.
\end{equation}
\end{lemma}

\textit{Proof}. It is sufficient to show that for all $k$ and $i$,
\begin{equation}
    \gamma_{k+1}\sum_{ij} (W_{ji}^{k+1})^2 = \gamma_{k}\sum_{ij} (W_{ij}^{k})^2.
\end{equation}

We prove by contradiction. Let $U^*, W^*$ be the global minimum of the loss function. Assuming that for an arbitrary $k$, 
\begin{equation}
    \gamma_{k+1}\sum_{ij} (W_{ji}^{*, k+1})^2 \ne \gamma_{k}\sum_{ij} (W_{ij}^{*, k})^2.
\end{equation}
Now, we introduce $W^a$ such that $W_{ji}^{a, k+1} = a W_{ji}^{*, k+1}$ and $W_{ji}^{a, k} = W_{ji}^{*, k} / a$. The loss without regularization is invariant under the transformation of $W^*\to W^a$, namely
\begin{equation}
    L_0(W^*) = L_0(W^a).
\end{equation}
In the regularization, all the terms remain invariant except two terms:
\begin{equation}
    \begin{cases}
        \gamma_{k+1}\sum_{ij} (W_{ji}^{*, k+1})^2\to \gamma_{k+1}\sum_{ij} (W_{ji}^{a, k+1})^2 = a^2 \gamma_{k+1}\sum_{ij} (W_{ji}^{*, k+1})^2\\
        \gamma_{k}\sum_{ij} (W_{ij}^{*, k})^2\to \gamma_{k}\sum_{ij} (W_{ji}^{a, k})^2 = a^{-2} \gamma_{k}\sum_{ij} (W_{ji}^{*, k})^2
    \end{cases}
\end{equation}
It could be shown that, the sum of $a^2 \gamma_{k+1}\sum_{ij} (W_{ji}^{*, k+1})^2$ and $a^{-2} \gamma_{k}\sum_{ij} (W_{ji}^{*, k})^2$ reaches its minimum when $a^2 = \sqrt{\frac{\gamma_{k}\sum_{ij} (W_{ji}^{*, k})^2}{\gamma_{k+1}\sum_{ij} (W_{ji}^{*, k+1})^2}}$. If $\gamma_{k+1}\sum_{ij} (W_{ji}^{*, k+1})^2 \ne \gamma_{k}\sum_{ij} (W_{ij}^{*, k})^2$, one can choose $a$ to minimize the regularization terms in the loss function such that $L(W^a) < L(W^*)$, indicating $W^*$ is not the global minimum. Thus, $\gamma_{k+1}\sum_{ij} (W_{ji}^{*, k+1})^2 \ne \gamma_{k}\sum_{ij} (W_{ij}^{*, k})^2$ cannot be true.
$\square$

\subsection{Proof of Proposition~\ref{prop: 0 is local minimum}}
\textit{Proof}. Let
\begin{equation}
    L_0 = \E_{\tilde{x}}  \E_{ \epsilon^2,...,\epsilon_{D}} \left(\sum_{i_1,i_2,...,i_D}^{d_1,d_2,...d_D} U_{i_D} \epsilon^{(D)}_{i_{D}}...\epsilon_{i_1}^{(1)} W_{i_1i}^{(1)} x_i  - y\right)^2.
\end{equation}
$L_0$ is a polynomial containing $2D+2$th order, $D+1$th order, and $0$th order terms in terms of parameters $U$ and $W$. The second order derivative of $L$ is thus a polynomial containing $2D$-th order and $(D - 1)$-th order terms; however, other orders are not possible. For $D\ge 2$, there are no constant terms in the Hessian of $L$, and there is at least a parameter in each of the terms. 

The Hessian of the full loss function with regularization is 
\begin{align}
    \frac{\partial^2 L}{\partial^2 U_i U_j}& = \frac{\partial^2 L_0}{\partial^2 U_i U_j} + (1-\delta_{ij})2\gamma_u(U_i + U_j) + \delta_{ij}2\gamma_u;\\
    \frac{\partial^2 L}{\partial^2 W^{(i)}_{jk} U_l}& = \frac{\partial^2 L_0}{\partial^2 W^{(i)}_{jk} U_l} + 2(\gamma_w W^{(i)}_{jk} + \gamma_u U_l );\\
    \frac{\partial^2 L}{\partial^2 W^{(i)}_{jk} W^{(l)}_{mn}}& = \frac{\partial^2 L_0}{\partial^2 W^{(i)}_{jk} W^{(l)}_{mn}} + (1 - \delta_{il}\delta_{jm}\delta_{kn})2\gamma_w( W^{(i)}_{jk} + W^{(l)}_{mn}) + \delta_{il}\delta_{jm}\delta_{kn}2\gamma_w.
\end{align}
For $U=0$, $W=0$, the Hessian of $L_0$ is $0$, since each term in $L_0$ contains at least a $U$ or a $W$. The Hessian of $L$ becomes
\begin{align}
    \left.\frac{\partial^2 L}{\partial^2 U_i U_j}\right|_{U,W=0}& = \delta_{ij}2\gamma_u;\\
    \left.\frac{\partial^2 L}{\partial^2 W^{(i)}_{jk} U_l}\right|_{U,W=0}& = 0;\\
    \left.\frac{\partial^2 L}{\partial^2 W^{(i)}_{jk} W^{(l)}_{mn}}\right|_{U,W=0}& = \delta_{il}\delta_{jm}\delta_{kn}2\gamma_w.
\end{align}
The Hessian of $L$ is a positive-definite matrix. Thus, $U=0$, $W=0$ is always a local minimum of the loss function $L$. $\square$

\subsection{Proof of Proposition~\ref{prop: general solution condition and property}}
We first apply Lemma~\ref{lemma 3} to determine the condition for the nontrivial solution to exist. In particular, the Lemma must hold for $W^{(2)}$ and $W^{(1)}$, which leads to the following condition:
\begin{equation}
    \label{eq:layer1}
    ||b^{D-1} d_0^{D-1} [ b^{2D} d_0^D (\sigma^2 + d_0)^D A_0 + \gamma]^{-1}\E[xy]||^2 = 1.
\end{equation}
Note that the left-hand side is a continuous function that tends to $0$ as $b\to \infty$. Therefore, it is sufficient to find the condition that guarantees that there exists $b$ such that the l.h.s. is larger than $1$. For any $b$, the l.h.s. is a monotonically decreasing function of any eigenvalue of $A_0$, and so the following two inequalities hold:
\begin{equation}\label{eq: sufficient condition}
    \begin{cases}
        ||b^{D-1}d_0^{D-1}( b^{2D} d_0^D (\sigma^2 + d_0)^D \sigma_x^2 + \gamma)^{-1}\E[xy]|| \leq ||b^{D-1}d_0^{D-1}( b^{2D} d_0^D (\sigma^2 + d_0)^D a_{\rm min} + \gamma)^{-1}\E[xy]||\\
        ||b^{D-1}d_0^{D-1}( b^{2D} d_0^D (\sigma^2 + d_0)^D \sigma_x^2 + \gamma)^{-1}\E[xy]|| \geq ||b^{D-1}d_0^{D-1}( b^{2D} d_0^D (\sigma^2 + d_0)^D a_{\rm max} + \gamma)^{-1}\E[xy]||.
    \end{cases}
\end{equation}
The second inequality implies that if 
\begin{equation}
    ||b^{D-1}d_0^{D-1}[b^{2D} d_0^D (\sigma^2 + d_0)^D a_{\rm max} + \gamma]^{-1}\E[xy]|| > 1,
\end{equation}
a nontrivial solution must exist. This condition is equivalent to the existence of a $b$ such that
\begin{equation}\label{eq: multilayer special polynomial}
    d_0^D (\sigma^2 + d_0)^D a_{\rm max} b^{2D} -||\E[xy]|| b^{D-1} d_0^{D-1} < -\gamma,
\end{equation}
which is a polynomial inequality that does not admit an explicit condition for $b$ for a general $D$. Since the l.h.s is a continuous function that increases to infinity as $b\to \infty$, one sufficient condition for \eqref{eq: multilayer special polynomial} to hold is that the minimizer of the l.h.s. is smaller than $\gamma$. 

Note that the left-hand side of Eq.~\eqref{eq: multilayer special polynomial} diverges to $\infty$ as $b\to \pm \infty$ and tends to zero as $b\to 0$. Moreover, Eq.~\eqref{eq: multilayer special polynomial} is lower-bounded and must have a nontrivial minimizer for some $b>0$ because the coefficient of the $b^{D-1}$ term is strictly negative. One can thus find its minimizer by taking derivative. In particular, the left-hand side is minimized when
\begin{equation}
    b^{D+1} =  \frac{(D-1)||\E[xy]||}{2D d_0(\sigma^2 + d_0)^D a_{\rm max}},
\end{equation}
and we can obtain the following sufficient condition for \eqref{eq: multilayer special polynomial} to be satisfiable, which, in turn, implies that \eqref{eq:layer1} is satisfiable:
\begin{equation}
    \frac{D+1}{2D}||\E[xy]||d_0^{D-1}\left(\frac{(D-1)||\E[xy]||}{2D d_0 (\sigma^2 +d_0)^D a_{\rm max}} \right)^{\frac{D-1}{D+1}} > \gamma,
\end{equation}
which is identical to the proposition statement in \eqref{eq: existence condition}.

%



Now, we come back to condition~\eqref{eq: sufficient condition} to derive a sufficient condition for the trivial solution to be the only solution. The first inequality in Condition~\eqref{eq: sufficient condition} implies that if 
\begin{equation}
    ||b^{D-1} d_0^{D-1} [ b^{2D} d_0^D (\sigma^2 + d_0)^D a_{\rm min} + \gamma]^{-1}\E[xy]|| \leq 1,
\end{equation}
the only possible solution is the trivial one, and the condition for this to hold can be found using the same procedure as above to be 
\begin{equation}
    \frac{D+1}{2D}||\E[xy]||d_0^{D-1}\left(\frac{(D-1)||\E[xy]||}{2D d_0 (\sigma^2 +d_0)^D a_{\rm min}} \right)^{\frac{D-1}{D+1}}\leq \gamma,
\end{equation}
which is identical to~\eqref{eq: nonexistence condition}. 

We now prove the upper bound in \eqref{eq: solution upper bound}. Because for any $b$, the first condition in \eqref{eq: sufficient condition} gives an upper bound, and so any $b$ that makes the upper bound less than $1$ cannot be a solution. This means that any $b$ for which
\begin{equation}
    ||b^{D-1}d_0^{D-1}[ b^{2D} d_0^D (\sigma^2 + d_0)^D a_{\rm min} + \gamma]^{-1}\E[xy]|| \leq 1
\end{equation}
cannot be a solution. This condition holds if and only if
\begin{equation}\label{eq: multilayer special polynomial 2}
    d_0^D (\sigma^2 + d_0)^D a_{\rm min} b^{2D} -||\E[xy]|| b^{D-1} d_0^{D-1} > -\gamma.
\end{equation}
Because $\gamma >0$, one sufficient condition to ensure this is that there exists $b$ such that 
\begin{equation}
    d_0 (\sigma^2 + d_0)^D a_{\rm min} b^{2D} -||\E[xy]|| b^{D-1} > 0,
\end{equation}
which is equivalent to
\begin{equation}
    b > \left[\frac{||\E[xy]||}{ d_0 (\sigma^2 + d_0)^D a_{\rm min}}\right]^{\frac{1}{D+1}}. 
\end{equation}
Namely, any solution $b^*$ satisfies
\begin{equation}
    b^* \leq \left[\frac{||\E[xy]||}{ d_0(\sigma^2 + d_0)^D a_{\rm min}}\right]^{\frac{1}{D+1}}.
\end{equation}
We can also find a lower bound for all possible solutions. When $D> 1$, another sufficient condition for Eq.~\eqref{eq: multilayer special polynomial 2} to hold is that there exists $b$ such that 
\begin{equation}
    ||\E[xy]|| d_0^{D-1} b^{D-1}  < \gamma.
\end{equation}
because the $b^{2D}$ term is always positive. This condition then implies that any solution must satisfy:
\begin{equation}
 b^* \geq \frac{1}{d_0}\left[\frac{\gamma}{||\E[xy]||}\right]^{\frac{1}{D-1}}.
\end{equation}
For $D=1$, we have by Theorem~\ref{theo: main theo two layer} that 
\begin{equation}
    b^*>0
\end{equation}
if and only if $\E[xy]> \gamma$. This means that 
\begin{equation}
    b^{*}\geq \lim_{\eta \to 0^+}\lim_{D\to 1^+}\frac{1}{d_0}\left[\frac{\gamma + \eta}{||\E[xy]||}\right]^{\frac{1}{D-1}} = \begin{cases}
        \infty & \text{if $\E[xy] \geq \gamma$};\\
        0 & \text{if $\E[xy]< \gamma$}.
    \end{cases}.
\end{equation}
This finishes the proof. $\square$

\subsection{Proof of Theorem~\ref{theo: general global minimum condition}}
\textit{Proof}. When nontrivial solutions exist, we are interested in identifying when $b=0$ is not the global minimum. To achieve this, we compare the loss of $b=0$ with the other solutions. Plug the trivial solution into the loss function in Eq.~\eqref{eq: arbitrary depth main loss}, the loss is easily identified to be $L_{\rm trivial}= E[y^2]$. 

For the nontrivial minimum, defining $f$ to be the model,
\begin{align}
    f(x) & := \sum_{i,i_1,i_2,...,i_D}^{d,d_1,d_2,...d_D} U_{i_D} \epsilon^{(D)}_{i_{D}}...\epsilon_{i_2}^{(2)} W_{i_2i_1}^{(2)}\epsilon_{i_1}^{(1)} W_{i_1i}^{(1)} x\\
    \label{eq:fx}
    & = \eta d_0^D b^{2D} \E[xy]^T [b^{2D} d_0^D(\sigma^2 + d_0)^D A_0 + \gamma I]^{-1} x,
\end{align}
where, similar to the previous proof, we have defined $\sum_{i_1,...,i_D}^{d_1,...d_D}  \epsilon^{(D)}_{i_{D}}...\epsilon_{i_1}^{(1)}:= \eta$ such that  $\E[\eta] = \prod_i^D d_i = d_0^D$ and $\E[\eta^2] = \prod_i^Dd_i(\sigma_i^2 + d_i) := d_0^D(\sigma^2 + d_0)^D$. With this notation, The loss function becomes
\begin{align}
    &\E_x\E_{\eta}(f(x) -y)^2 + L_2\ reg.\\
    &=  \E_{x,\eta}[f(x)^2] - 2\E_{x,\eta}[yf(x)] + \E_x[y^2] + L_2\ reg. \\
    & =  \sum_i \frac{d_0^{3D}(\sigma^2 + d_0)^D b^{4D} a_i \E[x'y]_i^2 }{[d_0^D (\sigma^2+d_0)^D a_i b^{2D} + \gamma]^2} - 2  \sum_i\frac{d_0^{2D} b^{2D} \E[x'y]^2_i}{d_0^D (\sigma^2+d_0)^D a_i b^{2D} + \gamma} + \E_x[y^2] +  L_2\ reg. 
    \label{eq:minimum loss}
\end{align}
The last equation is obtained by rotating $x$ using a orthogonal matrix such that $R^{-1}A_0R = \diag(a_i)$ and denoting the rotated $x$ as $x' = Rx$. With $x'$, The $L_2\ reg$ term takes the form of 
\begin{align}
\label{eq: L2 reg}
    L_2\ reg. &= \gamma D d_0^2 b^2 + \gamma \sum_i \frac{d_0^{2D} b^{2D}  ||\E[x'y]_i||^2}{(d_0^D (\sigma^2 + d_0)^D b^{2D} a_i + \gamma)^2}.
\end{align}

Combining the expressions of \eqref{eq: L2 reg} and \eqref{eq:minimum loss}, we obtain that the difference between the loss at the non-trivial solution and the loss at $0$ is
\begin{equation}
    -\sum_i \frac{d_0^{2D}b^{2D} \E[x'y]_i^2 }{[d_0^D (\sigma^2+d_0)^D a_i b^{2D} + \gamma]} +  \gamma D d_0^2 b^2.
\end{equation}
Satisfaction of the following relation thus guarantees that the global minimum is nontrivial:
\begin{equation}
\label{eq:condition non trivial}
   \sum_i \frac{d_0^{2D}b^{2D} \E[x'y]_i^2 }{[d_0^D (\sigma^2+d_0)^D a_i b^{2D} + \gamma]} \ge \gamma D d_0^2 b^2.
\end{equation}
This relation is satisfied if
\begin{align}
   \frac{d_0^{2D}b^{2D} ||\E[xy]||^2 }{[d_0^D (\sigma^2+d_0)^D a_{max} b^{2D} + \gamma]} & \ge \gamma D d_0^2 b^2\\
   \frac{b^{2D-2}}{[d_0^D (\sigma^2+d_0)^D a_{max} b^{2D} + \gamma]} & \ge \frac{\gamma D }{d_0^{2D - 2}||\E[xy]||^2}.
   \label{eq:condition non trivial relaxed}
\end{align}
The derivative or l.h.s. with respect to $b$ is
\begin{equation}
    \frac{b^{2D-3}[(2D-2)\gamma-2d_0^D (\sigma^2+d_0)^D a_{max}d^{2D}]}{[d_0^D (\sigma^2+d_0)^D a_{max} b^{2D} + \gamma]^2}.
\end{equation}
For $b, \gamma \in (0, \infty)$, the derivative dives below $0$, indicating the l.h.s. of \eqref{eq:condition non trivial relaxed} has a global maximum at a strictly positive $b$. The value of $b$ is found when setting the derivative to $0$, namely
\begin{align}
   \frac{b^{2D-3}[(2D-2)\gamma-2d_0^D (\sigma^2+d_0)^D a_{max}d^{2D}]}{[d_0^D (\sigma^2+d_0)^D a_{max} b^{2D} + \gamma]^2} & = 0\\
   (2D-2)\gamma-2d_0^D (\sigma^2+d_0)^D a_{max}d^{2D} & = 0 \\
   b^{2D} & = \frac{(D - 1) \gamma}{d_0^D (\sigma^2+d_0)^D a_{max}}.
\end{align}
The maximum value then takes the form
\begin{equation}
    \frac{(D-1)^\frac{D-1}{D}}{D\gamma^\frac{1}{D}d_0^{D-1}(\sigma^2 + d_0)^{D-1}a_{max}^\frac{D-1}{D}}.
\end{equation}
The following condition thus guarantees that the global minimum is non-trivial
\begin{align}
    \frac{(D-1)^\frac{D-1}{D}}{D\gamma^\frac{1}{D}d_0^{D-1}(\sigma^2 + d_0)^{D-1}a_{max}^\frac{D-1}{D}} & \ge \frac{\gamma D }{d_0^{2D - 2}||\E[xy]||^2}\\
    ||\E[xy]||^2 & \ge \frac{\gamma^\frac{D+1}{D} D^2 (\sigma^2 + d_0)^{D-1} a_{max}^\frac{D-1}{D}}{d_0^{D-1}(D-1)^\frac{D-1}{D}}.
\end{align}
This finishes the proof. $\square$

\subsection{Proof of Theorem~\ref{theo: variance}}

\textit{Proof}. The model prediction is:
\begin{align}
    f(x) & := \sum_{i,i_1,i_2,...,i_D}^{d,d_1,d_2,...d_D} U_{i_D} \epsilon^{(D)}_{i_{D}}...\epsilon_{i_2}^{(2)} W_{i_2i_1}^{(2)}\epsilon_{i_1}^{(1)} W_{i_1i}^{(1)} x\\
    & = \eta d_0^D b^{2D} \E[xy]^T [b^{2D} d_0^D(\sigma^2 + d_0)^D \sigma_x^2 I + \gamma I]^{-1} x.
\end{align}
One can find the expectation value and variance of a model prediction:
\begin{equation}
    \E_\eta [f(x)] = \frac{d_0^{2D} b^{2D} \E[xy]^T x}{b^{2D} d_0^D (\sigma^2 + d_0)^D \sigma_x^2 + \gamma}
\end{equation}
For the trivial solution, the theorem is trivially true. We thus focus on the case when the global minimum is nontrivial.

The variance of the model is 
\begin{align}
    \V[f(x)] & = \E[f(x)^2] - \E[f(x)]^2\\
    & = \frac{(\sigma^2 + d_0)^D d_0^{3D} b^{4D} (\E[xy]^Tx)^2}{[b^{2D} d_0^D (\sigma^2 + d_0)^D \sigma_x^2 + \gamma]^2} - \frac{ d_0^{4D} b^{4D} (\E[xy]^Tx)^2}{[b^{2D} d_0^D (\sigma^2 + d_0)^D]^2 \sigma_x^2 + \gamma]^2}\\
    & = \frac{ d_0^{3D}[(\sigma^2 + d_0)^D - d_0^D] b^{4D} (\E[xy]^Tx)^2}{[b^{2D} d_0^D (\sigma^2 + d_0)^D \sigma_x^2 + \gamma]^2}\\
    & =  \frac{ d_0^{3D}[(\sigma^2 + d_0)^D - d_0^D] b^{2D+2} (\E[xy]^Tx)^2}{||\E[xy]||^2},
\end{align}
where the last equation follows from Eq.~\eqref{eq: solution condition}. The variance can be upper-bounded by applying \eqref{eq: solution upper bound}, 
\begin{equation}
    \V[f(x)] \le \frac{ d_0^{D}[(\sigma^2 + d_0)^D - d_0^D] (\E[xy]^Tx)^2}{(\sigma^2 + d_0)^{2D}\sigma_x^2} \propto \frac{ d_0^{D}[(\sigma^2 + d_0)^D - d_0^D] }{(\sigma^2 + d_0)^{2D}}.
\end{equation}
We first consider the limit $d_0 \to \infty$ with fixed $\sigma^2$:
\begin{equation}
    \V[f(x)] \propto \frac{D d_0^{2D-1}\sigma^2}{(d_0 + \sigma^2)^{2D}}  = O\left(\frac{1}{d_0} \right).
\end{equation}
For the limit $\sigma^2\to \infty$ with $d_0$ fixed, we have
\begin{equation}
    \V[f(x)] = O \left( \frac{1}{(\sigma^2)^{D}} \right).
\end{equation}
Additionally, we can consider the limit when $D\to \infty$ as we fix both $\sigma^2$ and $d_0$:
\begin{equation}
    \V[f(x)] = O \left( e^{-D2\log[(\sigma^2 + d_0)/d_0]} \right),
\end{equation}
which is an exponential decay. $\square$

\section{Exact Form of $b^*$ for $D=1$}\label{app sec: exact form}

Note that our main result does not specify the exact value of $b^*$. This is because $b^*$ must satisfy the condition in Eq.~\eqref{eq:b main}, which is equivalent to a high-order polynomial in $b$ with coefficients being general functions of the eigenvalues of $A_0$, whose solutions are generally not analytical by Galois theory. One special case where an analytical formula exists for $b$ is when $A_0 =\sigma_x^2I$. Practically, this can be achieved for any (full-rank) dataset if we disentangle and rescale the data by the whitening transformation: $x\to\sigma_x\sqrt{A_0^{-1}} x$. In this case, we have
\begin{equation}\label{eq: two layer homogeneous solution}
    b_*^2 = \frac{\sqrt{\frac{\gamma_w}{\gamma_u}}||\E[xy]|| - \gamma_w}{(\sigma^2 +d_1)\sigma_x^2},
\end{equation}
and 
\begin{equation}
    v= \pm \sqrt{\frac{\sqrt{\frac{\gamma_u}{\gamma_w}}||\E[xy]|| - \gamma_u}{\sigma_x^2(\sigma^2+d_1)}}\frac{\E[xy]}{||\E[xy]||},
\end{equation}
where $v = W_{i:}$.

\section{Effect of Bias}
\label{app sec:bias}
This section studies a deep linear network with biases for every layer and compares it with the no-bias networks. We first study a general case when the data does not receive any preprocessing. We then show that the problem reduces to the setting we considered in the main text under the common data preprocessing schemes that centers the input and output data: $\E[x]=0$, and $\E[y] = 0$. 
\subsection{Two-layer network}
The two-layer linear network with bias is defined as
\begin{equation}
\label{eq:TwoLayerBias}
    f_b(x; U, W, \beta^U, \beta^W) = \sum_i \epsilon_i U_i (W_{i:} \cdot x + \beta^W_i) + \beta^U,
\end{equation}
where $\beta^W \in \mathbb{R}^{d_1}$ is the bias in the hidden layer, and $\beta^U \in\mathbb{R}$ is the bias at the output layer. The loss function is
\begin{align}
\label{eq:TwoLayerBiasLoss}
    L_b(U, W, \beta^U, \beta^W) = & \E_{\epsilon, x, y}\left[(\sum_i \epsilon_i U_i (W_{i:} \cdot x + \beta^W_i) + \beta^U - y)^2\right] + L_2\\
     = & \E_{x, y}\left[ (UWx + U\beta^W + \beta^U - y) ^ 2 + \sigma^2\sum_i U_i^2 (W_{i:}\cdot x + \beta^W_i)^2\right] \nonumber\\
     & + \gamma_u(||U||^2 + (\beta^U)^2) + \gamma_w(||W||^2 + ||\beta^W||^2).
\end{align}

It is helpful to concatenate $x$ and $1$ into a single vector $x' := (x, 1)^{\rm T}$ and concatenate $W$ and $\beta^W$ into a single matrix $W'$ such that $W$, $\beta^W$, $x$, and $W'$, $x'$ are related via the following equation
\begin{equation}
    Wx + \beta^W = W' x'. 
\end{equation}
Using $W'$ and $x'$, the model can be written as
\begin{equation}
    f_b(x', U, W', \beta^U) = \sum_i\epsilon_i U_i W'_{i:}\cdot x' + \beta^U.
\end{equation}
The loss function simplifies to
\begin{equation}
\label{eq:TwoLayerBiasLossconcatenate}
    L_b(U, W', \beta) = \E_{\epsilon, x, y}[(\sum_i \epsilon_i U_i W'_{i:} \cdot x' + \beta^U - y)^2] + \gamma_u (||U||^2 + (\beta^U)^2) + \gamma_w ||W'||^2.
\end{equation}
Note that \eqref{eq:TwoLayerBiasLossconcatenate} contains similar rescaling invariance between $U$ and $W'$ and the invariance of aligning $W'_{i:}$ and $W'_{j:}$. One can thus obtain the following two propositions that mirror Lemma \ref{lemma 1} and \ref{lemma 2}.
\begin{proposition}
\label{prop:TwoLayerBiasU}
    At the global minimum of \eqref{eq:TwoLayerBiasLoss}, $U_j^2 = \frac{\gamma_w}{\gamma_u} \left(\sum_i W_{ji}^2 + (\beta^W_j)^2\right)$.
\end{proposition}
\begin{proposition}
\label{prop:TwoLayerBiasW}
    At the global minimum, for all $i$ and $j$, we have
    \begin{equation}
        \begin{cases}
            U_i^2 = U_j^2;\\
            U_iW_{i:} = U_jW_{j:};\\
            U_i \beta^W_i = U_j \beta^W_j.\\
        \end{cases}
    \end{equation}
\end{proposition}
The proofs are omitted because they are the same as those of Lemma \ref{lemma 1} and \ref{lemma 2}, substituting $W$ by $W'$. 
Following a procedure similar to finding the solution for a no-bias network, one can prove the following theorem.
\begin{theorem}
\label{theo:TwoLayerBiasSolution}
    The global minimum of Eq.~\eqref{eq:TwoLayerBiasLoss} is of the form
    \begin{equation}
\label{eq:TwoLayerNonTrivialSolution}
    \begin{cases}
            U = b\mathbf{r};\\
            \beta^U = \frac{d_1b\left[(d_1 + \sigma^2)\frac{\gamma_u}{\gamma_w}b - 1\right]v\E[x] - \left(d_1\frac{\gamma_u}{\gamma_w}b^2 - 1\right)\E[y]}{(d_1 + \sigma^2)d_1\frac{\gamma_u^2}{\gamma_w^2}b^4 + (\gamma_u - 2)d_1\frac{\gamma_u}{\gamma_w}b^2 + \gamma_u + 1};\\
            W =\mathbf{r} b\left\{\E[x]\left[b\frac{\gamma_u}{\gamma_w}(d_1 + b\sigma^2) - 1\right]\beta^U + \E[xy]\right\}^T[b^2(d_1 + \sigma^2)A_0 + \gamma_w I]^{-1};\\
            \beta^W = -\mathbf{r}\frac{\gamma_u}{\gamma_w} b \beta^U,
        \end{cases}
\end{equation}
where $b$ satisfies
\begin{equation}
    \gamma_u b^2 = b^2\frac{\gamma_w (\frac{\E[y]S_1S_3}{S_4}\E[x] - \E[xy]) (M^{-1})^2 (\frac{\E[y]S_1S_3}{S_4}\E[x] - \E[xy])^T + \frac{\gamma_u^2}{\gamma_w} \left(\frac{S_3}{S_4}\E[y] - b\frac{S_2}{S_4}\E[x]M^{-1}\E[xy]\right)^2}{\left(b\frac{S_2S_1}{S_4}\E[x]M^{-1}\E[x]^T - 1\right)^2},
\end{equation}
where $M, S_1, S_2, S_3, S_4$ are functions of the model parameters and $b$, defined in Eq.~\eqref{app eq: term shorthands}.
\end{theorem}
\textit{Proof.} First of all, we derive a handy relation satisfied by $\beta^U$ and $\beta^W$ at all the stationary points. The zero-gradient condition of the stationary points gives
\begin{equation}
\begin{cases}
    \E_{x, y}[2(UWx + U\beta^W + \beta^U - y)]U + 2\gamma_w \beta^W = 0;\\
    \E_{x, y}[2(UWx + U\beta^W + \beta^U - y)] + 2\gamma_u \beta^U = 0,\\
\end{cases}
\end{equation}
leading to 
\begin{align}
    U \gamma_u \beta^U + \gamma_w \beta^W  = 0\\
    \beta^W_i  = -\frac{\gamma_u}{\gamma_w} U_i \beta^U.
\end{align}
Proposition \ref{prop:TwoLayerBiasU} and proposition \ref{prop:TwoLayerBiasW} implies that we can define $b := |U_i|$ and $bv := U_iW_{i:}$. Consequently, $U_i\beta^W_i = -\frac{\gamma_u}{\gamma_w} b^2 \beta^U$, and the loss function can be written as
\begin{align}
    \mathbb{E}_{x} \left[ \left(bd_1 \sum_{j}v_jx_j - (d_1\frac{\gamma_u}{\gamma_w}b^2 - 1)\beta^U - y \right)^2 +  b^2d_1 \sigma^2\left(\sum_{k} v_{k}x_k -\frac{\gamma_u}{\gamma_w} b \beta^U \right)^2\right] &+   \gamma_u d_1 b^2 \nonumber\\
    + \gamma_w d_1 ||v||_2^2 + \gamma_u \left(\frac{b^2d_1\gamma_u}{\gamma_w}+ 1\right)(\beta^U)^2.
\end{align}
The respective zero-gradient condition for $v$ and $\beta^U$ implies that for all stationary points, 
\begin{equation}
\begin{cases}
    v = [b^2(d_1 + \sigma^2)A_0 + \gamma_w I]^{-1} b\left\{\E[x]\left[b\frac{\gamma_u}{\gamma_w}(d_1 + b\sigma^2) - 1\right]\beta^U + \E[xy]\right\};\\
    \beta^U = \frac{d_1b\left[(d_1 + \sigma^2)\frac{\gamma_u}{\gamma_w}b - 1\right]v\E[x] - \left(d_1\frac{\gamma_u}{\gamma_w}b^2 - 1\right)\E[y]}{(d_1 + \sigma^2)d_1\frac{\gamma_u^2}{\gamma_w^2}b^4 + (\gamma_u - 2)d_1\frac{\gamma_u}{\gamma_w}b^2 + \gamma_u + 1}.
\end{cases}
\end{equation}
To shorten the expressions, we introduce
\begin{equation}\label{app eq: term shorthands}
\begin{cases}
    M = b^2(d_1 + \sigma^2)A_0 + \gamma_w I;\\
    S_1 = b\frac{\gamma_u}{\gamma_w}(d_1 + b\sigma^2) - 1;\\
    S_2 = d_1b\left[(d_1 + \sigma^2)\frac{\gamma_u}{\gamma_w}b - 1\right];\\
    S_3 = d_1\frac{\gamma_u}{\gamma_w}b^2 - 1;\\
    S_4 = (d_1 + \sigma^2)d_1\frac{\gamma_u^2}{\gamma_w^2}b^4 + (\gamma_u - 2)d_1\frac{\gamma_u}{\gamma_w}b^2 + \gamma_u + 1.
\end{cases}
\end{equation}
With $M, S_1, S_2, S_3, S_4$, we have
\begin{equation}
    \begin{cases}
        v = M^{-1} b(\E[x]S_1\beta^U + \E[xy]);\\
        \beta^U = \frac{S_2 v\E[x] - S_3 \E[y]}{S_4}.
    \end{cases}
\end{equation}

The inner product of $v$ and $\E[x]$ can be solved as
\begin{equation}
    v\E[x] = b\frac{\frac{S_3}{S_4}\E[x]M^{-1}\E[x]S_1\E[y] - \E[x]M^{-1}\E[xy]}{b\frac{S_2}{S_4}\E[x]M^{-1}\E[x]S_1 - 1}.
\end{equation}

Inserting the expression of $v\E[x]$ into the expression of $\beta^U$ one obtains
\begin{equation}
    \beta^U = \frac{S_3\E[y] - bS_2\E[x]M^{-1}\E[xy]}{b\E[x]M^{-1}\E[x]S_1S_2 - S_4}
\end{equation}
The global minimum must thus satisfy
\begin{align}
    \gamma_u b^2 & = \gamma_w ||v||^2 + \gamma_u \frac{b^2 d_1 \gamma_u}{\gamma_w} (\beta^U)^2\\
     & = b^2\frac{\gamma_w (\frac{\E[y]S_1S_3}{S_4}\E[x] - \E[xy]) (M^{-1})^2 (\frac{\E[y]S_1S_3}{S_4}\E[x] - \E[xy])^T + \frac{\gamma_u^2}{\gamma_w} \left(\frac{S_3}{S_4}\E[y] - b\frac{S_2}{S_4}\E[x]M^{-1}\E[xy]\right)^2}{\left(b\frac{S_2S_1}{S_4}\E[x]M^{-1}\E[x]^T - 1\right)^2}.
\end{align}
This completes the proof. $\square$

\begin{remark}
    As in the no-bias case, we have reduced the original problem to a one-dimensional problem. However, the condition for $b$ becomes so complicated that it is almost impossible to understand. That being said, the numerical simulations we have done all carry the bias terms, suggesting that even with the bias term, the mechanisms are qualitatively similar, and so the approach in the main text is justified.
\end{remark}

When $\E[x] = 0$, the solution can be simplified a little:
\begin{equation}
    \begin{cases}
        U = \textbf{r}b;\\
        \beta^U = - \frac{d_1\frac{\gamma_u}{\gamma_w}b^2 - 1}{(d_1 + \sigma^2)d_1\frac{\gamma_u^2}{\gamma_w^2}b^4 + (\gamma_u - 2)d_1\frac{\gamma_u}{\gamma_w}b^2 + \gamma_u + 1} \E[y];\\
        W = \textbf{r} b \E[xy]^T [b^2(d_1 + \sigma^2)A_0 + \gamma_w I]^{-1};\\
        \beta^W = \textbf{r} \frac{\gamma_u}{\gamma_w} b \frac{d_1\frac{\gamma_u}{\gamma_w}b^2 - 1}{(d_1 + \sigma^2)d_1\frac{\gamma_u^2}{\gamma_w^2}b^4 + (\gamma_u - 2)d_1\frac{\gamma_u}{\gamma_w}b^2 + \gamma_u + 1} \E[y],
    \end{cases}
\end{equation}
where the value of $b$ is either $0$ or determined by
\begin{equation}
    \gamma_u = \gamma_w |\E[xy]^T [b^2(d_1 + \sigma^2)A_0 + \gamma_w I]^{-1}|^2 + \frac{\gamma_u^2}{\gamma_w}\E[y]^2 \left(\frac{d_1\frac{\gamma_u}{\gamma_w}b^2 - 1}{(d_1 + \sigma^2)d_1\frac{\gamma_u^2}{\gamma_w^2}b^4 + (\gamma_u - 2)d_1\frac{\gamma_u}{\gamma_w}b^2 + \gamma_u + 1}\right)^2.
\end{equation}
In this case, the expression of $W$ is identical to the no-bias model. The bias of both layers is proportional to $\E[y]$. The equation determining the value of $b$ is also similar to the no-bias case. The only difference is the term proportional to $\E[y]^2$.

Lastly, the solution becomes significantly simplified when both $\E[x] =0$ and $\E[y] = 0$. When this is the case, the solution reverts to the case when there is no bias. In practice, it is a common and usually recommended practice to subtract the average of $x$ and $y$ from the data and achieve precisely $\E[x]=0$ and $\E[y] = 0$. We generalize this result to deeper networks in the next section. 

\subsection{Deep linear network}
Let $\beta$ be a $\left(\sum_i^D d_i + 1\right)$-dimensional vector concatenating all $\beta^{(1)}, \beta^{(2)}, ..., \beta^{(D)}, \beta^U$, and denoting the collection of all the weights $U$, $W^{(D)}$, ..., $W^{(1)}$ by $w$, the model of a deep linear network with bias is defined as
\newcommand{\bias}{\rm{bias}}
\begin{align}
\label{eq:DeepLinearBias}
    & f_b(x, W^{(D)}, ..., W^{(1)}, U, \beta^{(D)}, ..., \beta^{(1)}, \beta^U)\\
    = & U (\epsilon^{(D)} \circ (W^{(D)} (... (\epsilon^{(2)}  \circ  (W^{(2)} (\epsilon^{(1)} \circ (W^{(1)} x + \beta^{(1)})) + \beta^{(2)})) ...) + \beta^D)) + \beta^U\\
    \nonumber = & U (\epsilon^{(D)} \circ (W^{(D)} (... (\epsilon^{(2)}  \circ  (W^{(2)} (\epsilon^{(1)} \circ (W^{(1)} x))))))) \\ \nonumber & + U (\epsilon^{(D)} \circ (W^{(D)} (... (\epsilon^{(2)}  \circ  (W^{(2)} (\epsilon^{(1)} \circ \beta^{(1)}))))))\\ & + U (\epsilon^{(D)} \circ (W^{(D)} (... (\epsilon^{(2)}  \circ  \beta^{(2)})))) + ... + U (\epsilon^{(D)} \circ \beta^{(D)}) + \beta^U\\
    = & U (\epsilon^{(D)} \circ (W^{(D)} (... (\epsilon^{(2)}  \circ  (W^{(2)} (\epsilon^{(1)} \circ (W^{(1)} x))))))) + \bias(w, \beta),
\end{align}
where
\begin{align}
    \begin{split}
    {\rm bias}(w, \beta) = & U (\epsilon^{(D)} \circ (W^{(D)} (... (\epsilon^{(2)}  \circ  (W^{(2)} (\epsilon^{(1)} \circ \beta^{(1)}))))))\\ & + U (\epsilon^{(D)} \circ (W^{(D)} (... (\epsilon^{(2)}  \circ  \beta^{(2)})))) + ... + U (\epsilon^{(D)} \circ \beta^{(D)}) + \beta^U,
    \end{split}
\end{align}
and $\circ$ denotes Hadamard product. The loss function is
\begin{equation}
\label{eq:DeepLinearBiasLoss}
    L_b(x, y, w, \beta) = \E_{\epsilon, x, y}[(f_b(x, w, \beta) - y)^2] + L_2(w, \beta).
\end{equation}
Proposition \ref{prop:TwoLayerBiasU} and Proposition \ref{prop:TwoLayerBiasW} can be generated to deep linear network. Similar to the no-bias case, we can reduce the landscape to a $1$-dimensional problem by performing induction on $D$ and using the $2$-dimensional case as the base step. However, we do not solve this case explicitly here because the involved expressions now become too long and complicated even to write down, nor can they directly offer too much insight. We thus only focus on the case when the data has been properly preprocessed. Namely, $\E[x] = 0$ and $\E[y] = 0$.

For simplicity, we assume that the regularization strength for all the layers are identically $\gamma$. The following theorem shows that When $\E[x] = 0$ and $\E[y] = 0$, the biases vanish for an arbitrarily deep linear network: 
\begin{theorem}
    Let $\E[x] = 0$ and $\E[y] = 0$. The global minima of Eq.~\eqref{eq:DeepLinearBiasLoss} have $\beta^{(1)} = 0, \beta^{(2)} = 0, ..., \beta^{(D)} = 0, \beta^U = 0$.
\end{theorem}
\textit{Proof.}
At the global minimum, the gradient of the loss function vanishes. In particular, the derivatives with respect to $\beta$ vanish:
\begin{align}
\label{eq:MultiLayerDeriv}
    \frac{\partial L_b(x, y, w, \beta)}{\partial \beta_i} & = 0;\\
    \E_{\epsilon, x, y}\left[\frac{\partial f_b(x, w, \beta)}{\partial \beta_i}(f_b(x, w, \beta) - y)\right] + \gamma\beta_i& = 0;\\
    \E_{\epsilon, x, y}\left[\frac{\partial \bias(w, \beta)}{\partial \beta_i}(f_b(x, w, \beta) - y)\right] + \gamma\beta_i & = 0;\\
    \E_{\epsilon}\left[\frac{\partial \bias(w, \beta)}{\partial \beta_i}(f_b(\E[x], w, \beta) - \E[y])\right] + \gamma\beta_i & = 0,
    \label{eq:MultiLayerDerivLast}
\end{align}
where $\beta_i$ is the $i$th element of $\beta$.
The last equation is obtained since $f_b(x, w, \beta)$ is a linear function of $x$. Using the condition $\E[x] = 0 $ and $ \E[y] = 0$, Equation \eqref{eq:MultiLayerDerivLast} becomes
\begin{equation}
    \E_{\epsilon}\left[\frac{\partial \bias(w, \beta)}{\partial \beta_i} \bias(w, \beta)\right] + \gamma\beta_i = 0.
    \label{equ:bias_derivative}
\end{equation}
$\bias(w, \beta)$ is a linear combination of $\beta_i$. Consequently, $\partial \bias(w, \beta)/\partial \beta_i$ does not depend on $\beta$, and $\bias(w, \beta) \partial \bias(w, \beta)/\partial \beta_i$ is a linear combination of $\beta_i$. {Furthermore, the linearity of $\bias(w, \beta)$ indicates that $\beta \cdot \partial \bias(w, \beta)/\partial \beta = \bias(w, \beta)$. The solution to the equation \eqref{equ:bias_derivative} also takes the form of 
\begin{equation}
    \beta^* = \arg\min_{\beta} (||\bias(w, \beta)||^2 + \gamma ||\beta||^2).
    \label{equ:bias_min}
\end{equation}
The only choice of $\beta$ that minimizes both the first term and the second term in \eqref{equ:bias_min} is, regardless of the value of $w$,
\begin{equation}
    \beta = 0.
\end{equation}
}
This finishes the proof. $\square$

Thus, for a deep linear network, a model without bias is good enough to describe data satisfying $\E[x] = 0$ and $\E[y] = 0$, which could be achieved by subtracting the mean of the data.

\end{document}